\documentclass[10pt,twocolumn,letterpaper]{article}

\usepackage{cvpr}
\usepackage{times}
\usepackage{epsfig}
\usepackage{graphicx}
\usepackage{amsmath}
\usepackage{amssymb}
\usepackage{booktabs}
\usepackage{multirow}
\usepackage{subcaption}
\usepackage{bm}
\usepackage{xcolor}
\usepackage{authblk}
\usepackage{capt-of}
\usepackage[heightadjust=all]{floatrow}

\usepackage{tikz}
\usepackage{pgfplots}
\pgfplotsset{compat=1.14}
\usepgfplotslibrary{groupplots}
\usepackage[export]{adjustbox}
\usetikzlibrary{positioning, arrows.meta, backgrounds}

\definecolor{main1}{RGB}{230, 30, 100} 
\definecolor{main2}{RGB}{0, 140, 210} 
\definecolor{main3}{RGB}{240, 170, 0} 
\definecolor{main4}{RGB}{67, 176, 71} 
\definecolor{main5}{RGB}{136, 86, 167} 

\DeclareMathOperator*{\argmin}{arg\,min}

\usepackage[pagebackref=true,breaklinks=true,letterpaper=true,colorlinks,bookmarks=false]{hyperref}

\cvprfinalcopy 

\def\cvprPaperID{7805} 

\ifcvprfinal\fi
\begin{document}
\newfloatcommand{capbtabbox}{table}[][\FBwidth]
\title{Local Implicit Grid Representations for 3D Scenes}

\makeatletter
\renewcommand\Authfont{\fontsize{11.5}{14.4}\selectfont}
\renewcommand\AB@affilsepx{\qquad \protect\Affilfont}
\makeatother
\author[1,2]{Chiyu ``Max" Jiang}
\author[2]{Avneesh Sud}
\author[2]{Ameesh Makadia}
\author[2,3]{Jingwei Huang}
\author[4]{\\ Matthias Nie{\ss}ner}
\author[2]{Thomas Funkhouser}
\affil[1]{UC Berkeley}
\affil[2]{Google Research}
\affil[3]{Stanford University}
\affil[4]{Technical University of Munich}
\renewcommand*{\Authsep}{ \ }
\renewcommand*{\Authands}{ \ }


\setlength{\abovedisplayskip}{0.25\abovedisplayskip}
\setlength{\belowdisplayskip}{0.25\belowdisplayskip}
\setlength{\abovecaptionskip}{0.5\abovecaptionskip}
\setlength{\belowcaptionskip}{0.25\belowcaptionskip}

\twocolumn[{%
\renewcommand\twocolumn[1][]{#1}%
\maketitle
\vspace{-4em}
\begin{center}
    \centering
    \resizebox{\textwidth}{!}{
\begin{tikzpicture}[node distance=0em, auto]

\tikzset{>=latex}

\node (tsne) {\includegraphics[height=5cm,trim={0 0 0 0},clip]{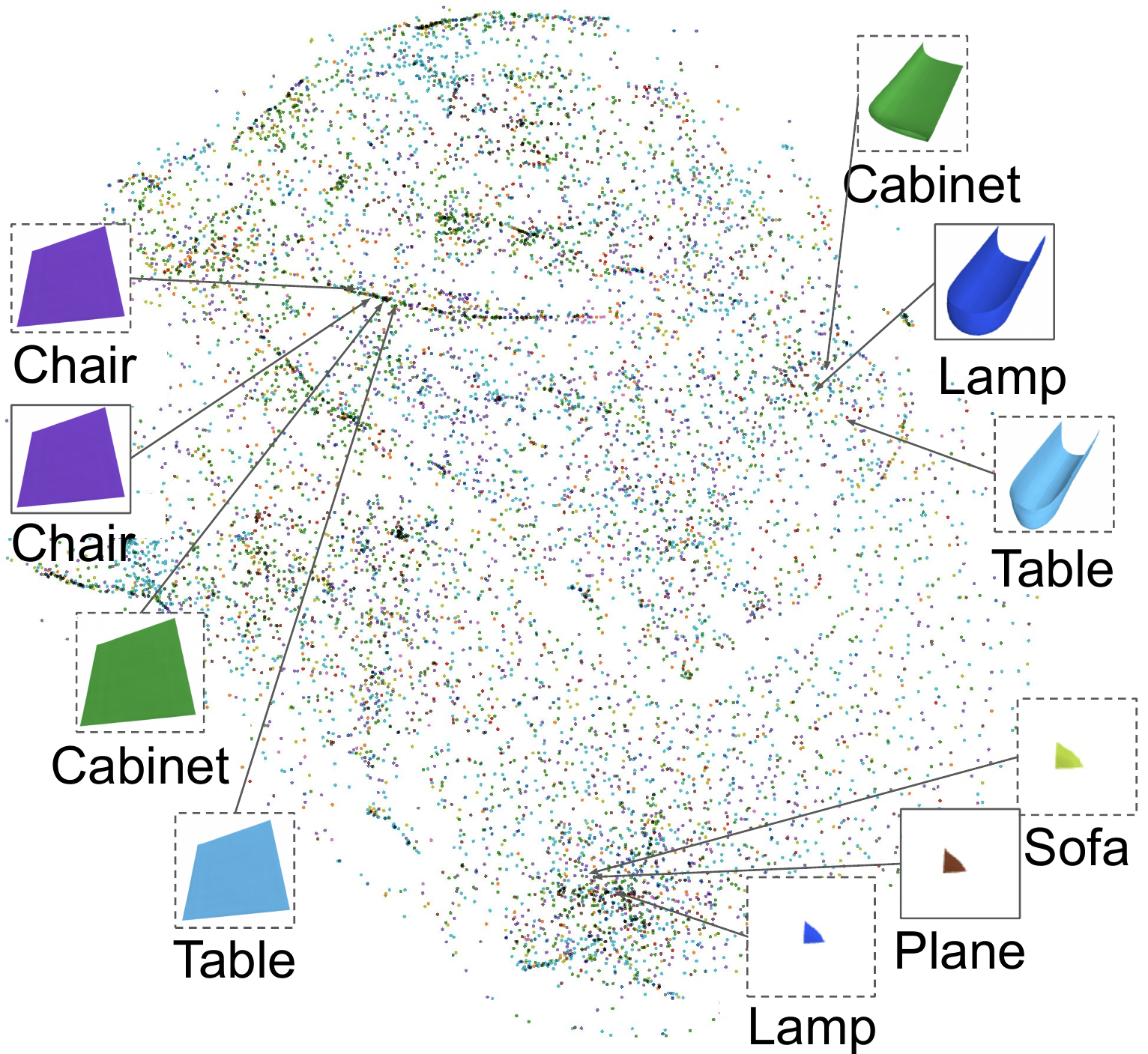}}; 

\node [right=0em of tsne] (recon) {\includegraphics[height=5cm,trim={0 0 0 0},clip]{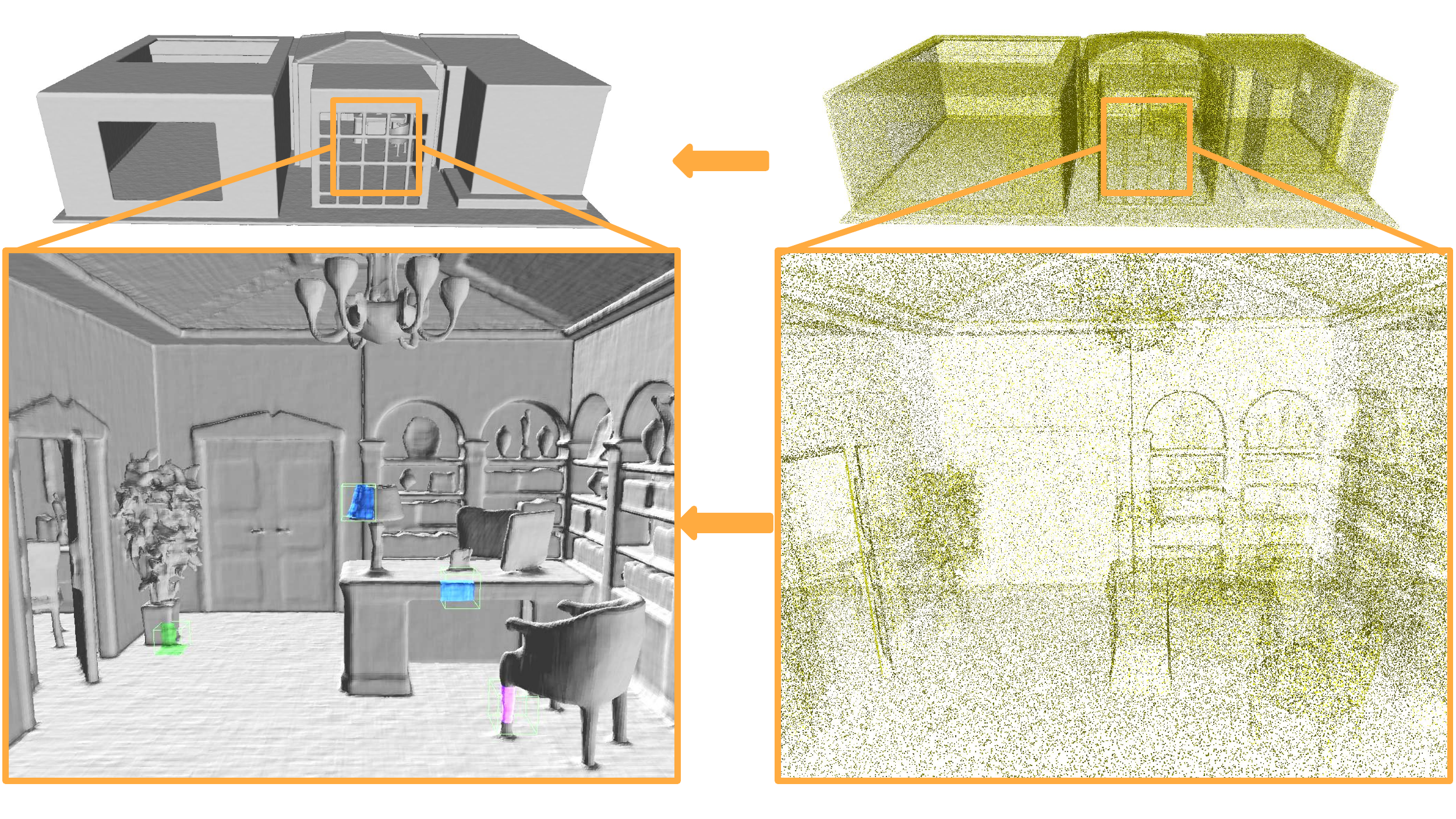}}; 

\node [left=0em of tsne] (part) {\includegraphics[height=5cm,trim={0 0 0 0},clip]{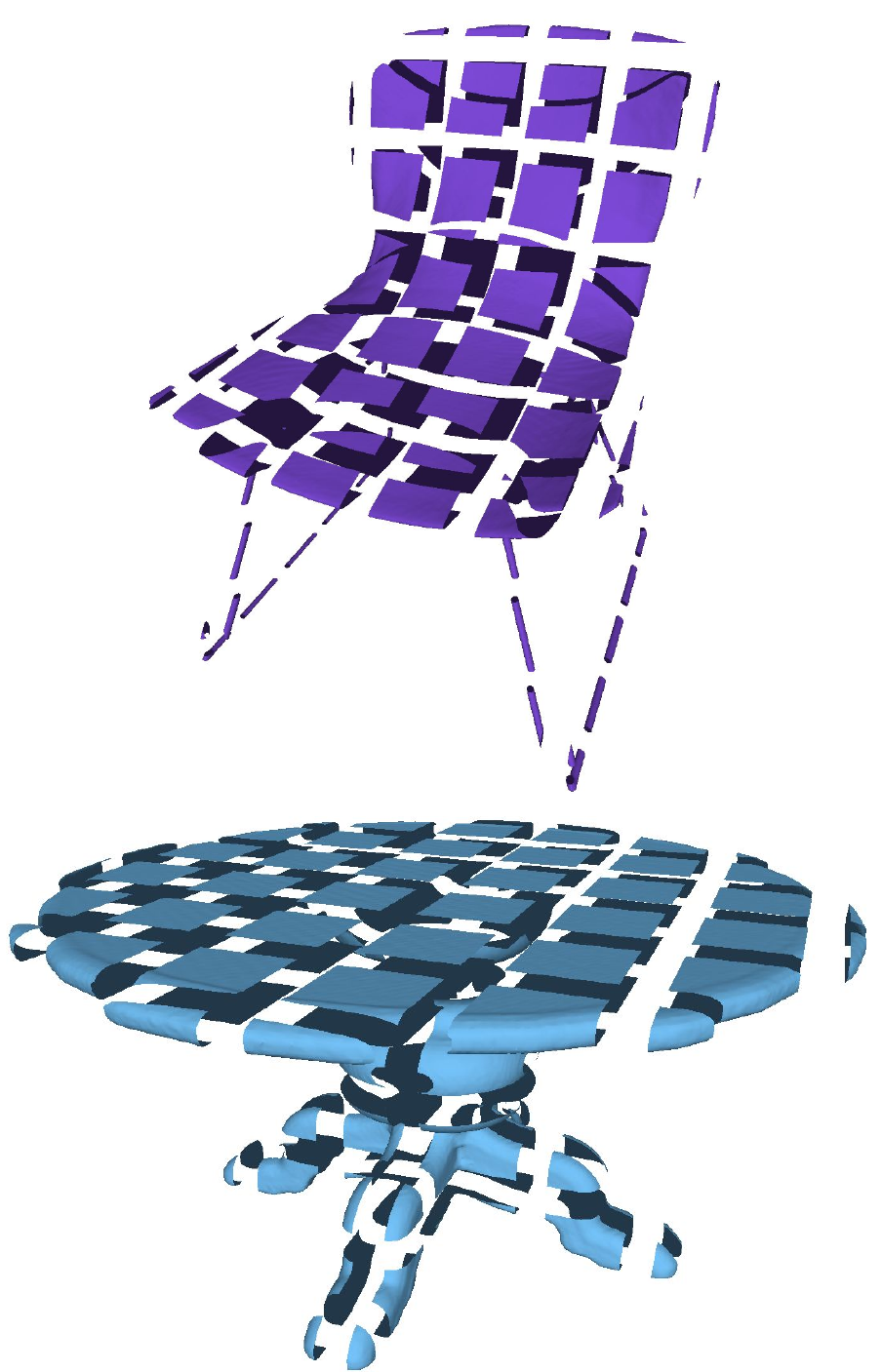}}; 

\node [below=0em of part] (cap0) {(a) Training parts from ShapeNet.};
\node [below=0em of tsne, align=left, shift={(1.5em, 0)}] (cap1) {(b) t-SNE plot of part embeddings.};
\node [below=0em of recon, align=left] (cap2) {(c) Reconstructing entire scenes with Local Implicit Grids};
\draw [color=gray, ->] (5, -0.6) -- (1, -0.3);
\end{tikzpicture}
}

    \captionof{figure}{\small We learn an embedding of parts from objects in ShapeNet \cite{chang2015shapenet} using a part autoencoder with an implicit decoder. We show that this representation of parts is generalizable across object categories, and easily scalable to large scenes. By localizing implicit functions in a grid, we are able to reconstruct entire scenes from points via optimization of the latent grid.}
    \label{fig:overview}
\end{center}%
\vspace{1em}
}]
\begin{abstract}
Shape priors learned from data are commonly used to reconstruct 3D objects from partial or noisy data.   Yet no such shape priors are available for indoor scenes, since typical 3D autoencoders cannot handle their scale, complexity, or diversity.   In this paper, we introduce Local Implicit Grid Representations, a new 3D shape representation designed for scalability and generality.  The motivating idea is that most 3D surfaces share geometric details at some scale -- i.e., at a scale smaller than an entire object and larger than a small patch.   We train an autoencoder to learn an embedding of local crops of 3D shapes at that size.   Then, we use the decoder as a component in a shape optimization that solves for a set of latent codes on a regular grid of overlapping crops such that an interpolation of the decoded local shapes matches a partial or noisy observation.  We demonstrate the value of this proposed approach for 3D surface reconstruction from sparse point observations, showing significantly better results than alternative approaches.
\vfill
\end{abstract}

\section{Introduction}
Geometric representation for scenes has been central to various tasks in computer vision and graphics, including geometric reconstruction, compression, and higher-level tasks such as scene understanding, object detection and segmentation. An effective representation should generalize well across a wide range of semantic categories, scale efficiently to large scenes, exhibit a rich expressive capacity for representing sharp features and complex topologies, and at the same time leverage learned geometric priors acquired from data.

In the last years, several works have proposed new network architectures to allow conventional geometric representations such as point clouds \cite{qi2017pointnet, fan2017point, yang2019pointflow}, meshes \cite{wang2018pixel2mesh, groueix2018atlasnet}, and voxel grids \cite{dai2017shape, wu2016learning} to leverage data priors. More recently, a neural implicit representation \cite{chen2019learning,mescheder2019occupancy, park2019deepsdf} has been proposed as an alternative to these approaches for its expressive capacity for representing fine geometric details. However, the aforementioned works focus on learning representations for whole objects within one or a few categories, and they have not been studied in the context of generalizing to other categories, or scaling to large scenes.

In this paper we propose a learned 3D shape representation that generalizes and scales to arbitrary scenes. Our key observation is that although different shapes across different categories and scenes have vastly different geometric forms and topologies on a global scale, they share similar features at a certain local scale. For instance, sofa seats and car windshields have a similar curved parts, tabletops and airplane wings both have thin sharp edges, etc.. While no two shapes are the same at the macro scale, and all shapes on a micro-scale can be locally approximated by an angled plane, there exists an intermediate scale (a ``part scale''), where a meaningful shared abstraction for all geometries can be learned by a single deep neural network.   We aim to learn shape priors at that scale and then leverage them in a scalable and general 3D reconstruction algorithm.

To this end, we propose the Local Implicit Grid (LIG) representation, a regular grid of overlapping part-sized local regions, each encoded with an implicit feature vector.  We learn to encode/decode geometric parts of objects at a part scale by training an implicit function autoencoder on 13 object categories from ShapeNet \cite{chang2015shapenet}.  Then, armed with the pretrained decoder, we propose a mechanism to optimize for the Latent Implicit Grid representation that matches a partial or noisy scene observation.  Our representation includes a novel overlapping latent grid mechanism for confidence-weighted interpolation of learned local features for seamlessly representing large scenes. We illustrate the effectiveness of this approach by targeting the challenging application of scene reconstruction from sparse point samples, where we are able to faithfully reconstruct entire scenes given only sparse point samples and shape features learned from ShapeNet objects. Such an approach requires no training on  scene level data, where data is costly to acquire. We achieve  significant improvement both visually and quantitatively in comparison to state-of-the-art reconstruction algorithms for the scene reconstruction from point samples task (Poisson Surface Reconstruction \cite{kazhdan2006poisson,kazhdan2013screened}, or PSR, among other methods).

In summary, the main contributions of this work are:
\begin{itemize}
    \item We propose the Local Implicit Grid representation for geometry, where we learn and leverage geometric features on a part level, and associated methods such as the overlapping latent grid mechanism and latent grid optimization methods for representing and reconstructing scenes at high fidelity.
    \item We illustrate the significantly improved generalizability of our part-based approach in comparison to related methods that learn priors for entire objects -- i.e., we can reconstruct shapes from novel object classes after training only on chairs, or construct entire scenes after training only on ShapeNet parts.
    \item We apply our novel shape representation approach towards the challenging task of scene reconstruction from sparse point samples, and show significant improvement over the state-of-the-art approach (For Matterport reconstruction from 100/$m^2$ input points, an F-Score of 0.889 versus 0.455.
\end{itemize}

\section{Related Work}
\subsection{Geometric representation for objects}
In computer vision and graphics, geometric representations such as simplicial complexes (point clouds, line meshes, triangular meshes, tetrahedral meshes) have long been used for representing geometries for its flexibility and compactness. In recent years, various neural architectures have been proposed for analyzing or generating such representations. For instance for \cite{qi2017pointnet, wang2019dynamic} have been proposed for analyzing point cloud representations, and \cite{fan2017point, yang2019pointflow} for generating point clouds. \cite{masci2015geodesic, hanocka2019meshcnn, chiyu2019spherical, huang2019texturenet} have been proposed for analyzing signals on meshes, and \cite{wang2018pixel2mesh, groueix2018atlasnet, dai2019scan2mesh} for generating mesh representations. \cite{jiang2019ddsl} proposed a general framework for analyzing arbitrary simplicial complex based geometric signals. Naturally paired with 3D Convolutional Neural Networks (CNNs), voxel grids have also been extensively used as a 3D representation \cite{wu20153d, dai2018scancomplete, choy20163d}.

More recently, alternative representations have been proposed in the context of shape generation. Most related to our method are \cite{mescheder2019occupancy,park2019deepsdf,chen2019learning}, where the implicit surfaces of geometries are represented as spatial functions using fully-connected neural networks. Continuous spatial coordinates are fed as input features to the network which directly produces the values of the implicit functions, however these methods encode the entire shape using a global latent code. \cite{sitzmann2019scene} used such implicit networks to represent neural features instead of occupancies that can be combined with a differentiable ray marching algorithm to produce neural renderings of objects. Rather than learning a single global implicit network to represent the entire shape, \cite{saito2019pifu} learns a continuous per-pixel occupancy and color representation using implicit networks. Other novel geometric representations in the context of shape reconstruction include Structured Implicit Functions that serves as learned local shape templates \cite{genova2019learning}, and CvxNet \cite{deng2019cvxnets} which represents space as a convex combination of half-planes that are localized in space. These methods represent entire shapes using a single global latent vector, which can be decoded into continuous outputs with the associated implicit networks.

\subsection{Localized geometric representations}
Though using a single global latent code to represent entire geometries and scenes is appealing for its simplicity, it fails to capture localized details, and scales poorly to large scenes with increased complexities. \cite{xu2019disn} proposes to address the localization problem in the context of image to 3D reconstruction by first estimating a camera pose for the images followed by the projection of local 2D features to be concatenated with global latents for decoding. However, the scalability of such hybrid representations beyond single objects has yet to be shown. Similar to our approach, \cite{williams2019deep} uses a local patch based representation. However it is not trained on any data, hence is not able to leverage any shape priors from 3d datasets.  \cite{pauly2005example} combines shape patches extracted directly from a set of examples, which limits the shape expressibility.  Similar to our spatial partitioning of geometries into part grids, \cite{tang2018real} uses PCA-based decomposition to learn a reduced representation of geometric parts within TSDF grids of a fixed scale for the application of real-time geometry compression.  These methods do not support scalable reconstruction with learned deep implicit functions.

\subsection{Scene-level geometry reconstruction}
Most deep learning studies have investigated object reconstruction, with input either as an RGB/D image \cite{choy20163d, wang2018pixel2mesh,mescheder2019occupancy,chen2019learning,fan2017point,deng2019cvxnets,genova2019learning} or 3D points \cite{park2019deepsdf, liao2018deep, jiang2019convolutional}, and yet few have considered learning to reconstruct full scenes.  Scene level geometry reconstruction is a much more challenging task in comparison to single objects.  \cite{song2017semantic} performs semantic scene completion within the frustum of a single depth image. \cite{dai2018scancomplete} uses a 3D convolutional network with a coarse-to-fine inference strategy to directly regress gridded Truncated Signed Distance Function (TSDF) outputs from incomplete input TSDF. \cite{avetisyan2019scan2cad} tackles the scene reconstruction problem by CAD model retrieval, which produces attractive surfaces, at the expense of geometric inaccuracies. However, all of the methods require training on reliable and high-quality scene data. Though several real and synthetic scene datasets exist, such as SunCG \cite{song2016ssc}, SceneNet \cite{handa2016understanding}, Matterport3D \cite{Matterport3D}, and ScanNet \cite{dai2017scannet}, they are domain-specific and acquiring data for new scenes can be costly. In contrast to methods above that require training on scene dataset, our method naturally generalizes shape priors learned from object datasets and does not require additional training on scenes.

\section{Methods}
\subsection{Method overview}
We present a schematic overview of our method in Figure \ref{fig:overview}. We first learn an embedding of shape parts at a fixed scale from objects in a synthetic dataset using \textit{part autoencoders} (see Sec. \ref{ssec:partae}).  We show two interesting properties of such a latent embedding: (1) objects that originated from different categories share similar part geometries, validating the generalizability of such learned representations, and (2) parts that are similar in shape are close in the latent space. In order to scale to scenes of arbitrary sizes, we introduce an overlapping gridded representation that can layout these local representations in a scene (Sec. \ref{ssec:lig}). Using such part embeddings that can be continuously decoded spatially using a local implicit network, we are able to faithfully reconstruct geometries from only sparse oriented point samples by searching for a corresponding latent code using gradient descent-based optimization to match given observations (Sec. \ref{ssec:optim}), thus efficiently leveraging geometric priors learned from parts from the ShapeNet dataset.

\subsection{Learning a latent embedding for parts}\label{ssec:partae}
\paragraph{Data} Our part embedding model is learned from a collection of 20 million object parts culled from 3D-R$^2$N$^2$~\cite{choy20163d}, a 13-class subset of ShapeNet. As preprocessing, we normalize watertight meshes (generated with tools from \cite{mescheder2019occupancy}) into a $[0, 1]$ unit cube, leaving a margin of 0.1 at each side. To maintain the fidelity of the parts, we compute a signed distance function (SDF) at a grid resolution of $256^3$. Starting from the origin and with a stride of 16, all $32^3$ patches that have at least one point within $3/255$ of the shape surface are extracted as parts for training.

\paragraph{Part Autoencoder}
\begin{figure}
    \centering
    \includegraphics[width=\linewidth,trim={12em 2em 2em 2em},clip]{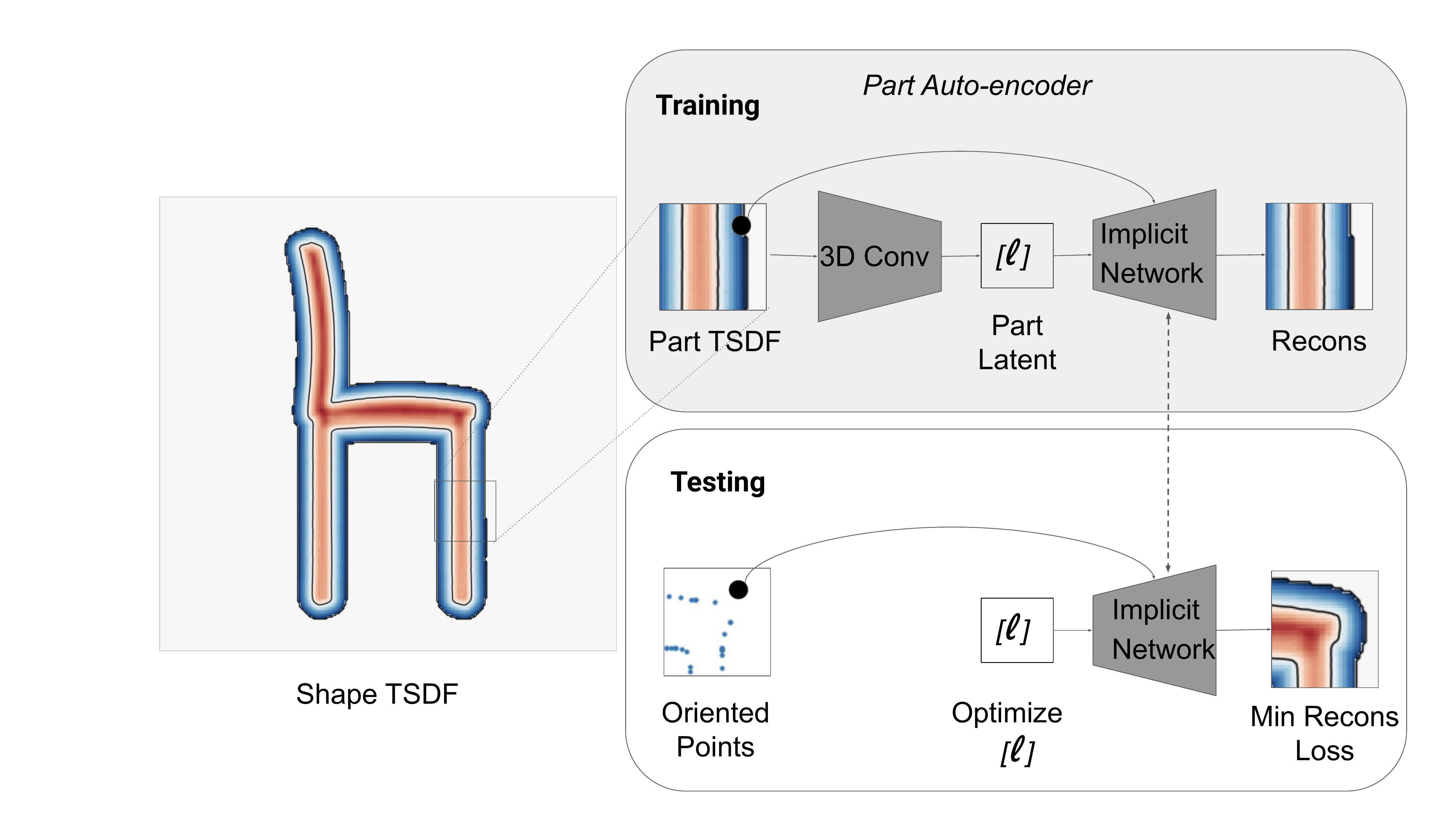}
    \caption{A schematic of the part autoencoder. At train time, crops of the TSDF grid from the ShapeNet dataset are used to train a part autoencoder, with a 3D CNN encoder and implicit network decoder. Interior and exterior points are sampled to supervise the network during training. At inference time, the pre-trained implicit network is attached to a Local Implicit Grid, and the corresponding latent values are optimized via gradient descent on observed interior/exterior points.}\label{fig:partae}
\end{figure}
We use a 3D CNN decorated with residual blocks for encoding such local TSDF grids, and a reduced IM-NET \cite{chen2019learning} decoder for reconstructing the part (See Fig. \ref{fig:partae}). An IM-NET decoder is a simple fully connected neural network with internal skip connections that takes in a latent code concatenated with a 3D point coordinate, and outputs the corresponding implicit function value at the point.
We train the network using point samples with binary in/out labels so that the network learns a continuous decision boundary of the binary classifier as the encoded surface. Since decoding a part is a much more simplified task than decoding an entire shape, we reduce the number of feature channels in each hidden layer of IM-NET by 4 fold, obtaining a leaner and more efficient decoder. To acquire a compact latent representation of parts, we further reduce the number of latent channels for each part to 32. We train the part autoencoder with 2048 random point samples that we sample from the SDF grid on-the-fly during training, where we sample points farther from the boundary with Gaussian-decaying probabilities. The sign of the sample points is interpolated from the sign of the original SDF grid. Furthermore, we truncate the input SDF grids to a value of $3/255$ and renormalize the grid to $[0, 1]$ for stronger gradients near the boundary.

We train the part autoencoder with binary cross entropy loss on the point samples, with an additional latent regularization loss to constrain the latent space of the learned embeddings. The loss is given as:
\begin{align}
    \mathcal{L}(\theta_{e}, \theta_{d}) =& \frac{1}{|\mathcal{P}||\mathcal{B}|}\sum_{i\in\mathcal{P}}\sum_{j\in\mathcal{B}} \mathcal{L}_{c}(D_{\theta_d}(\bm{x}_{i,j}, E_{\theta_e}(g_i)), \text{sign}(\bm{x}_{i,j})) \nonumber\\
    & + \lambda ||E_{\theta_e}(g_i)||_2
\end{align}
where $\mathcal{P}$ is the set of all training parts in a given mini-batch, $\mathcal{B}$ is the set of point samples sampled per part, $\mathcal{L}_c(\cdot, \cdot)$ is the binary cross-entropy loss with logits, $E_{\theta_e}$ is the convolutional encoder parameterized by trainable parameters $\theta_e$, $D_{\theta_d}$ is the implicit decoder parameterized by trainable parameters $\theta_d$, and $g_i$ is the input tsdf grid for the $i$-th part, $\text{sign}(\cdot)$ takes the sign of the corresponding point ${x}_{i,j}$.

\subsection{Local implicit grids}\label{ssec:lig}
\begin{figure}[t!]
    \centering
    \includegraphics[width=0.9\linewidth,trim={0 8em 0 7em},clip]{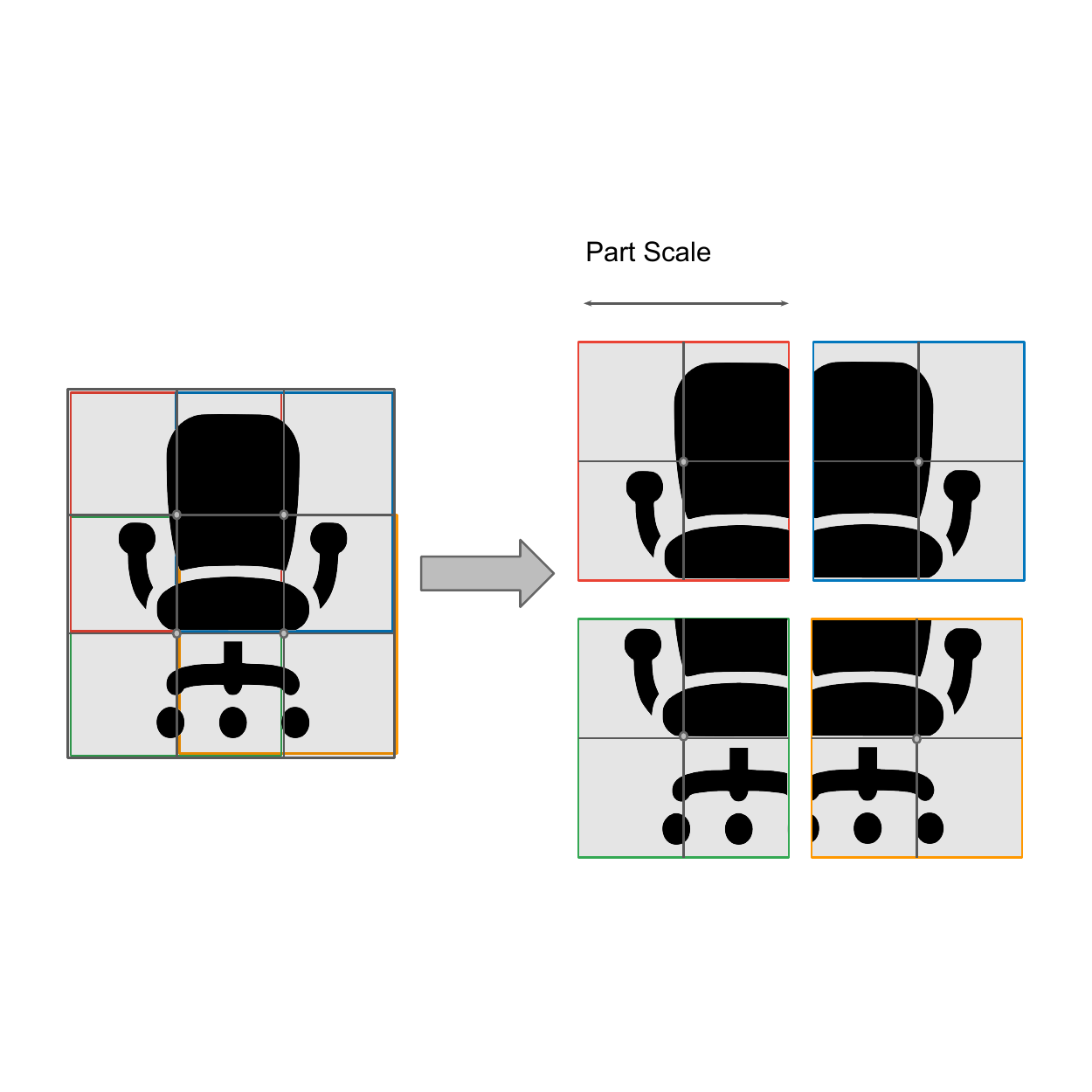}
    \caption{2D schematic for representing geometries with overlapping latent grids. The implicit value at any point is an bilinear/trilinear interpolation of implicit values acquired by querying 4/8 (2D/3D) neighbors with respect to each cell center.}
    \label{fig:olg}
\end{figure}
In order to use the learned part representations for representing entire objects and scenes, we lay out a sparse latent grid structure, where within each local grid cell the surface is continuously decoded from the local latent codes within the cell. In world coordinates, when querying for the implicit function value at location $\bm{x}$ against a single voxel grid cell centered at $\bm{x_{i}}$, the implicit value is decoded as:
\begin{align}
    f(\bm{x}, \bm{c}_i) = D_{\theta_d}(\bm{c}_{i}, \frac{2}{s}(\bm{x}-\bm{x}_{i}))
\end{align}
where $\bm{c}_{i}$ is the latent code corresponding to the part in cell $i$, and $s$ is the part scale. The coordinates are first being transformed into normalized local coordinates within the cell to $[-1, 1]$, before being queried against the decoder.

Though directly partitioning space into a voxel grid with latent channels within each cell gives decent performance, there will be discontinuities across voxel boundaries. Hence we propose the overlapping latent grid scheme, where each grid cell for a part overlaps with its neighboring cells by half the part scale (see Fig. \ref{fig:olg}). When querying for the implicit function value at an arbitrary position $\bm{x}$ against overlapping latent grids, the value is computed as a trilinear interpolation of independent queries to all cells that overlap at this position, which is 4 in 2 dimensions and 8 in 3 dimensions:
\begin{align}
    f(\bm{x}, \{\bm{c}_j | j \in \mathcal{N}\}) = \sum_{j\in\mathcal{N}}w_j D_{\theta_d}(\bm{c}_j, \frac{2}{s}(\bm{x}-\bm{x}_j))
\end{align}
where $\mathcal{N}_j$ is the set of all neighboring cells of point $\bm{x}$, and $w_j$ is the trilinear interpolation weight corresponding to cell $j$. Under such an interpolation scheme, the overall function represented by the implicit grid is guaranteed to be $C^0$ continuous. Higher-order continuity could be similarly acquired with higher degrees of polynomial interpolations, though we do not explore it in the scope of this study. For additional efficiency, since most grid cells do not have any points that fall into them, we use a sparse data structure for storing latent grid values, optimization, and decoding for the reconstructed surface, where empty space is assumed to be exterior space.

\subsection{Geometric encoding via latent optimization}\label{ssec:optim}
At inference time, when presented with a sparse point cloud of interior/exterior samples as input, we decompose space into a coarse grid and then perform optimization for the latent vectors associated with the grid cells in order to minimize the cost function for classifying sampled interior/exterior points. The initial values within the latent grid is initialized as random normal with a standard deviation of $10^{-2}$. If we denote the set of effective latent grid cells as $\mathcal{G}$, the corresponding latent code in each grid cell $c_j$, and the set of all sampled interior/exterior input points as $\mathcal{B}$, we optimize the latent codes for the minimal classification loss on the sampled points:
\begin{align}
    \argmin_{\bm{c}\in\mathcal{G}} \sum_{i\in\mathcal{B}}\sum_{j\in\mathcal{N}_i}&\mathcal{L}_c(f(\bm{x_i}, \{\bm{c}_j | j \in \mathcal{N}\}), \text{sign}(\bm{x}_i))+\lambda||\bm{c}_j||_2\label{eqn:loss}
\end{align}

\begin{figure}[t]
    \centering
    \includegraphics[width=0.8\linewidth]{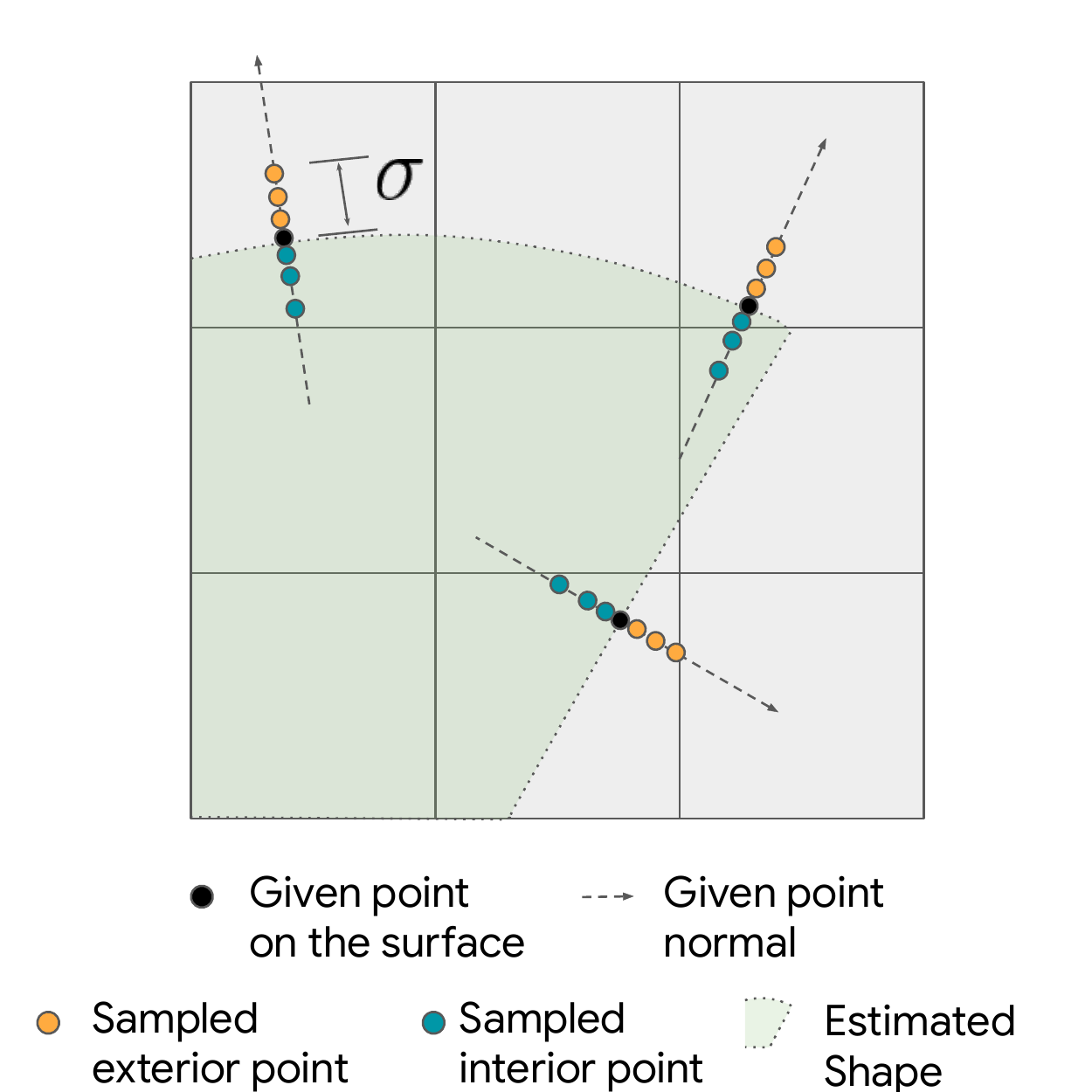}
    \caption{Schematic for reconstructing shapes based on sparse oriented point samples. Given original point samples on the surface with normals, we randomly sample $k$ samples along both sides of each normal vector and assign signs for these samples accordingly. The points are sampled with a Gaussian falloff probability, with a given standard deviation $\sigma$. The latent codes within the overlapping latent grids are updated via optimization for minimizing classification loss as in Eqn. \ref{eqn:loss}. The surface of the shape is reconstructed by densely querying the latent grid and extracting the zero-contour of the output logits.}
    \label{fig:optim}
\end{figure}

How do we acquire the signed point samples for performing this latent grid optimization? For autoencoding a geometry with a latent grid, the signed point samples are densely sampled near the surface of the given shape to be encoded. However, for the application of recovering surface geometry from sparse oriented point samples, we randomly sample interior and exterior points for each point sample along the given normal direction, with a Gaussian falloff probability parameterized by a standard deviation of $\sigma$. See Fig. \ref{fig:optim} for details. All grid cells that do not contain any point from the input point cloud is assumed to be an empty exterior volume. This is effective and works well for scenes that do not contain large enclosed volumes, but creates an artificial back-face in the enclosed interior. We detail a simple postprocessing algorithm to remove such artifacts resulting from the exterior empty space assumption. We provide more details about the additional postprocessing algorithm in the Appendix.

As our method requires optimizing over the learned latent space, it is reasonable to wonder if alternate models such as a variational autoencoder~\cite{kingma2014vae} or autodecoder~\cite{park2019deepsdf} would be a more appropriate choice, as both formulations incorporate a latent distribution prior. However, \cite{park2019deepsdf} observed the stochastic nature of the VAE made training difficult. Also, the autodecoder is fundamentally unable to scale to large numbers of parts at training as it requires fast storage and random access to all latent embeddings during training. These concerns motivated our decision to adopt an autoencoder formulation with a regularization loss to constrain the latent space.
\section{Experiments}
We ran a series of experiments to test the proposed LIG method.  We focus on two properties of our method: the generalization of our learned part representation, and the scalability of our learned shape representation to large scenes.  Our target application is reconstructing scenes from a sparse set of oriented point samples, a challenging task that requires learned part priors for detailed and accurate reconstruction.

\paragraph{Metrics} In all of our experiments, we evaluate geometric reconstruction quality with Chamfer Distance (CD), Normal Alignment (Normal), and F-Score. For Chamfer Distance and Normal Alignment, we base our implementation on \cite{mescheder2019occupancy} with small differences.
For object-level autoencoding experiments, we follow \cite{fan2017point, mescheder2019occupancy} and normalize the unit distance to be 1/10 of the maximal edge length of the current object’s bounding box. We estimate CD and Normal Alignment using 100,000 randomly sampled points on the ground truth and reconstructed meshes. For the two scene-level experiments, we randomly sample 2 million points on each mesh when estimating CD and Normal Alignment. When evaluating scene reconstructions, we use world coordinate scales (meters) for computing CD, since data is provided in a physically-meaningful scale. Additionally, in all experiments, we compute the F-Score at a threshold of $\tau$, as F-Score is a metric less sensitive to outliers. F-Score is the harmonic mean of recall (percentage of reconstruction to target distances under $\tau$) and precision (vice versa). For object reconstruction (Sec. \ref{ssec:partae}) we use $\tau=0.1$ and for scene reconstruction, we use $\tau=0.025$ (i.e., 2.5cm).

\begin{table}[]
\footnotesize
\resizebox{\columnwidth}{!}{
\begin{tabular}{@{}llp{3em}p{3em}lp{3em}p{4em}@{}}
\toprule
\multirow{2}{*}{Category}        & \multicolumn{3}{c}{IM-NET}                     & \multicolumn{3}{c}{Ours}                            \\ \cmidrule(l){2-7}
                                 & CD ($\downarrow$)   & Normal ($\uparrow$) & \multicolumn{1}{l|}{F-Score ($\uparrow$)} & CD ($\downarrow$)             & Normal ($\uparrow$)      & F-Score ($\uparrow$)  \\ \midrule
\multicolumn{1}{l|}{chair}       & 0.181 & 0.820  & \multicolumn{1}{l|}{0.505}  & \textbf{0.099} & \textbf{0.920}  & \textbf{0.710} \\ \midrule
\multicolumn{1}{l|}{airplane}    & 0.698 & 0.550 & \multicolumn{1}{l|}{0.151}  & 0.150          & 0.817          & 0.564          \\
\multicolumn{1}{l|}{bench}       & 0.229 & 0.719 & \multicolumn{1}{l|}{0.433}  & 0.054          & 0.905          & 0.857          \\
\multicolumn{1}{l|}{cabinet}     & 0.343 & 0.700 & \multicolumn{1}{l|}{0.230}  & 0.118          & 0.948          & 0.733          \\
\multicolumn{1}{l|}{car}         & 0.354 & 0.646 & \multicolumn{1}{l|}{0.240}   & 0.152          & 0.825          & 0.472          \\
\multicolumn{1}{l|}{display}     & 0.601 & 0.574 & \multicolumn{1}{l|}{0.130}  & 0.170          & 0.926          & 0.551          \\
\multicolumn{1}{l|}{lamp}        & 0.836 & 0.592 & \multicolumn{1}{l|}{0.120}   & 0.114          & 0.882          & 0.624           \\
\multicolumn{1}{l|}{loudspeaker} & 0.377 & 0.702 & \multicolumn{1}{l|}{0.246}  & 0.139          & 0.937          & 0.711          \\
\multicolumn{1}{l|}{rifle}       & 0.902 & 0.400 & \multicolumn{1}{l|}{0.080}  & 0.113          & 0.824          & 0.693          \\
\multicolumn{1}{l|}{sofa}        & 0.199  & 0.812 & \multicolumn{1}{l|}{0.484}   & 0.077          & 0.944          & 0.822          \\
\multicolumn{1}{l|}{table}       & 0.425 & 0.681  & \multicolumn{1}{l|}{0.242}  & 0.066          & 0.936          & 0.844          \\
\multicolumn{1}{l|}{telephone}   & 0.623 & 0.547 & \multicolumn{1}{l|}{0.120}    & 0.037          & 0.984          & 0.962          \\
\multicolumn{1}{l|}{vessel}      & 0.591 & 0.574 & \multicolumn{1}{l|}{0.147}  & 0.178          & 0.847          & 0.467          \\ \midrule
mean*                            & 0.435  & 0.666 & 0.274                       & \textbf{0.114} & \textbf{0.898} & \textbf{0.692} \\ \bottomrule
\end{tabular}
}
\caption{Shape autoencoding for autoencoders trained on only chairs and evaluated on all 13 categories. The mean corresponds to class-averaged mean of all out-of-training object categories.}\label{tab:exp1}
\end{table}

\subsection{Generalization of learned part representation}\label{ssec:ae}
\paragraph{Task} In order to investigate the generalization of the learned embedding by reducing the scale of the learned shape from object scale to part scale, we construct an investigative experiment of training the models to learn a shape autoencoder on a single category of objects (in this case, chairs in the training set of ShapeNet), and reconstructing examples from the all 13 object categories, including the other 12 unseen categories.

\paragraph{Baseline} As our main objective is to explore the gain in generalizability from learning an embedding of part scales, we benchmark our method against the original IM-NET decoder with a similar 3D convolution based encoder as the encoder part of our part autoencoder. To implement autoencoding for our method, we train our autoencoder on all the parts we extract from the training split of the chair category in ShapeNet. We then ``encode" the geometries of the unseen shapes using the latent optimization method that is described in Sec. \ref{ssec:optim}.

\begin{figure}[t!]
    \centering
    \includegraphics[width=.9\linewidth,trim={5em 0 3em 0},clip]{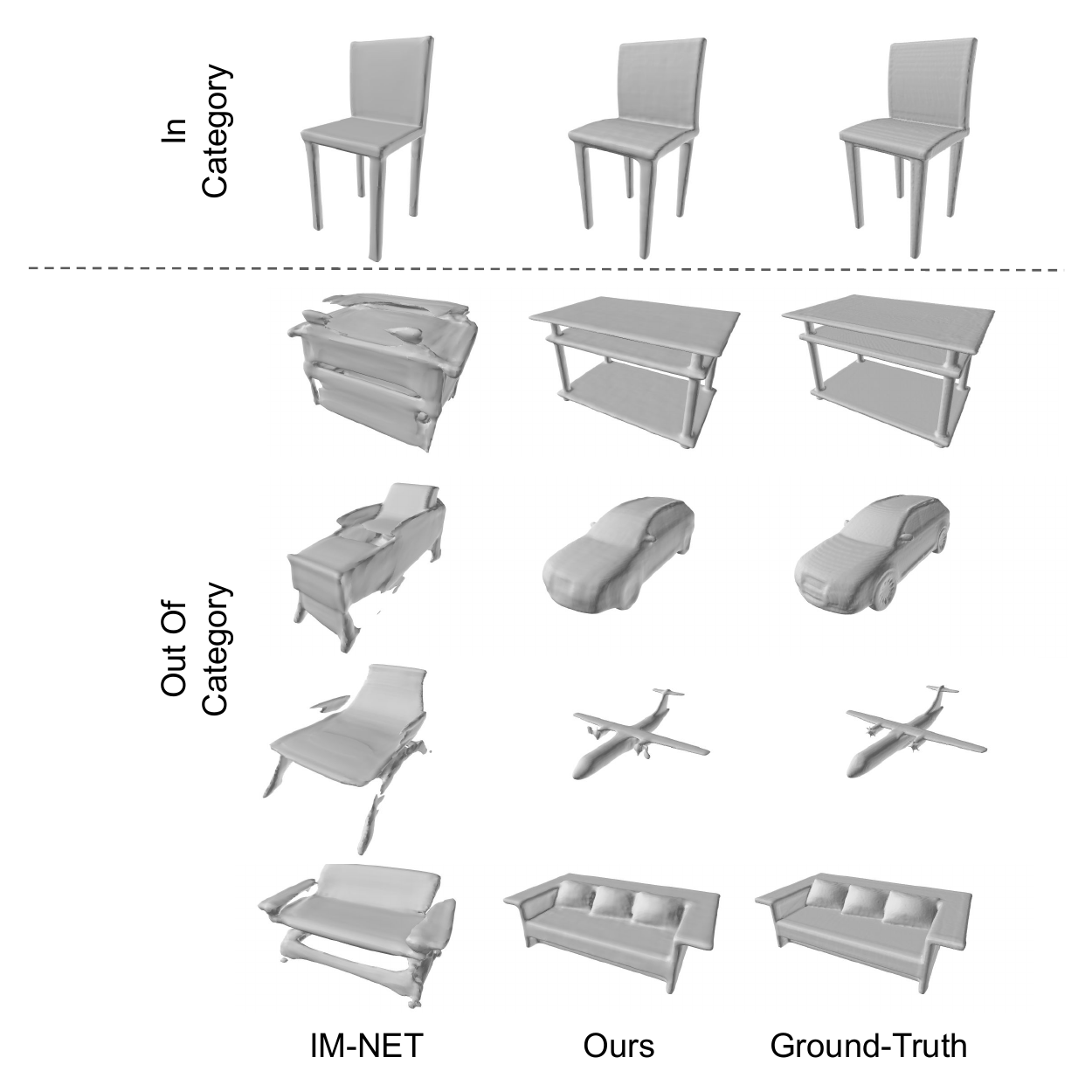}
    \caption{Qualitative comparison of autoencoded shape from in-category (chair) and out-of-category shapes. IM-NET trained to learn embeddings of one object category does not transfer well to unseen categories, while the part embedding learned by our local implicit networks is much more transferable across unseen categories.}\label{fig:exp1}
\end{figure}

\paragraph{Results Discussion} We quantitatively and qualitatively compare reconstruction performances in Table \ref{tab:exp1} and Figure \ref{fig:exp1}, respectively. Given an IM-NET that is trained to learn a latent representation of objects (in this scenario, chairs), the learned representation does not generalize to classes beyond the source class. Visually, IM-NET achieves good reconstructions on the source class as well as related classes (e.g., sofa), but performs poorly on semantically different classes (e.g., airplane). In contrast, the part representation learned by our local implicit networks is transferable across drastically different object categories.

\begin{figure}[t]
    \centering
    \includegraphics[width=\linewidth]{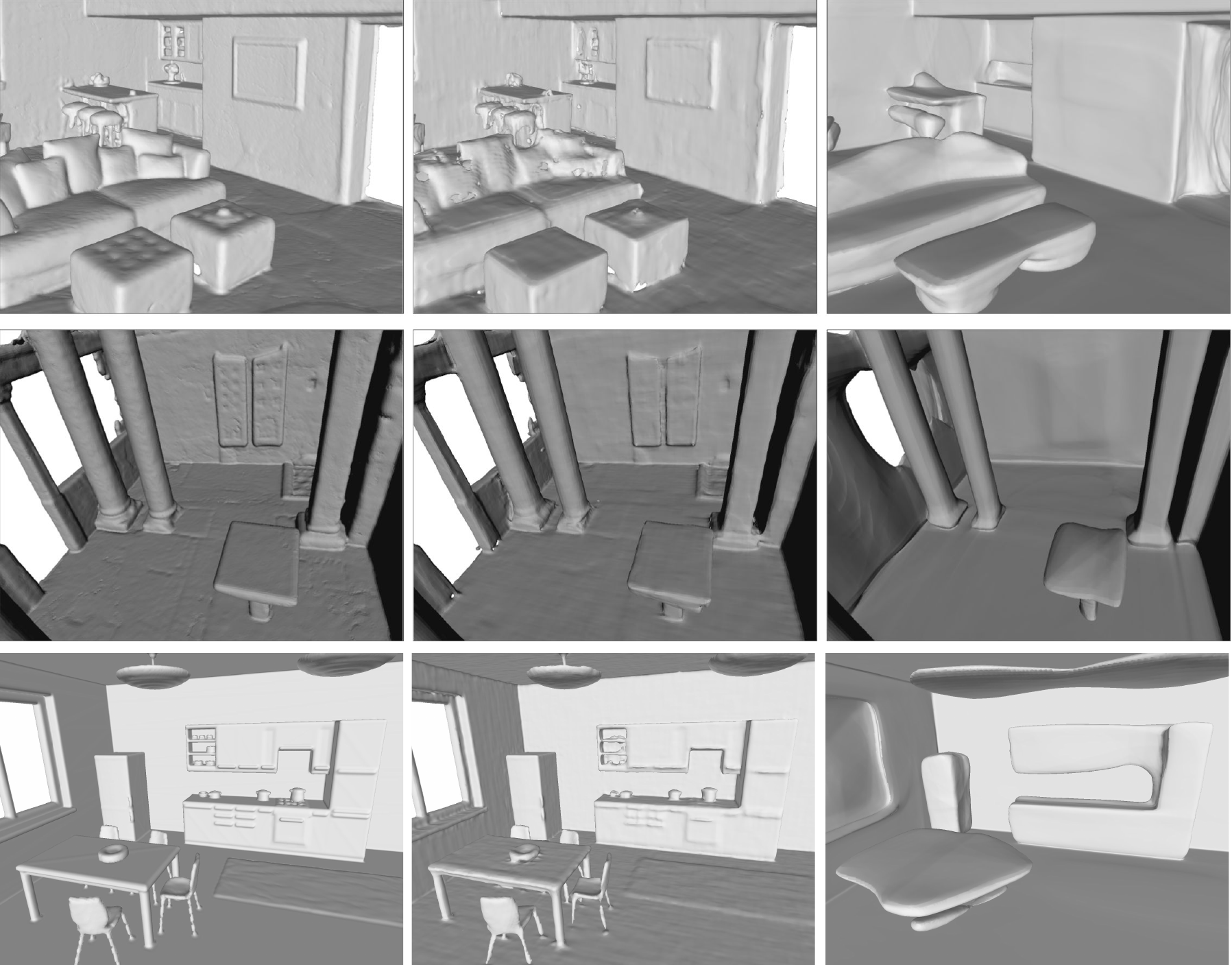}
    \caption{Qualitative comparison of the scene representational performance: Left to right: Ground truth scene, our reconstruction using sampling density 500 points/$m^2$, and IM-NET. First two rows from Matterport, last row from SceneNet.}\label{fig:exp2}

\end{figure}

\begin{table}[t!]
\small
\centering
\begin{tabular}{@{}llll@{}}
\toprule
Metrics & CD($\downarrow$)     & Normal($\uparrow$) & F-Score($\uparrow$) \\ \midrule
IM-NET  & 0.183 & 0.827 & 0.647  \\
Ours    & \textbf{0.007}  & \textbf{0.945} & \textbf{0.985}  \\ \bottomrule
\end{tabular}
\caption{Qualitative comparison of scene representational performance for IM-NET versus our method.}\label{tab:exp2}
\end{table}

\subsection{Scalability of scene representational power}\label{ssec:sca}
\paragraph{Task} As a second experiment, we investigate the increased representational power and scalability that we gain from learning a part-based shape embedding. The definition of the task is: given one scene, what is the best reconstruction performance we can get from either representation for memorizing and overfitting to the scene.

\begin{figure*}[t!]
    \centering
    \input{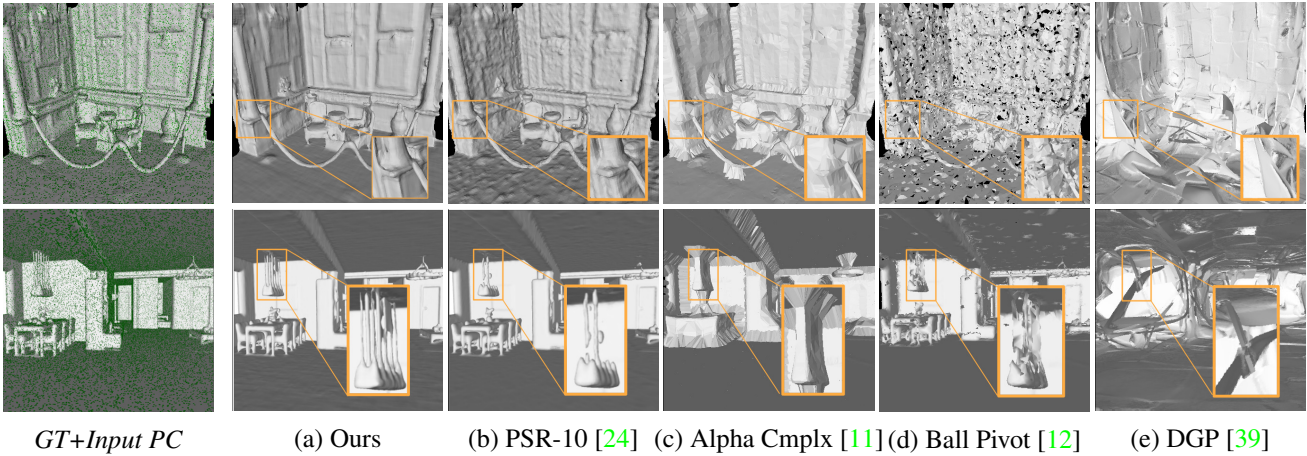}
    \caption{Qualitative comparisons of scene reconstruction performance from sparse oriented point samples. Our method is significantly better at reconstructing scenes from sparse point clouds compared to baseline methods, especially with respect to sharp edges and thin structures.}\label{fig:exp3_sc}
\end{figure*}

\paragraph{Baseline} Similar to the previous experiment, we compare directly with IM-NET for representational capacity towards a scene, as it is the decoder backbone that our method is based on, to investigate the improvement in scalability that we are able to gain by distributing geometric information in spatially localized grid cells versus a single global representation. For this task, as the objective is to encode one scene, we use the encoderless version of IM-NET, where during training time, the decoder only receives spatial coordinates of point samples (not concatenated with a latent code) that are paired with the signs of these points. For our method, we use latent optimization against the pretrained decoder for encoding the scenes, using 100k surface point samples from the scene, with a sampling factor of $k=10$ per point along the normal direction.

\paragraph{Data} We evaluate the representational qualities of the two methods on the meshes from the validation set of the Matterport 3D \cite{Matterport3D} scene dataset.  We perform the evaluations at the region level of the dataset, requiring the models to encode one region at a time. Additionally, we provide one example from SceneNet for visual comparison in Fig. \ref{fig:exp2}.

\paragraph{Results Discussion} The quantitative (Table \ref{tab:exp2}) and qualitative (Fig. \ref{fig:exp2}) results are presented. While IM-NET is able to reconstruct the general structure of indoor scenes such as smooth walls and floors, it fails to capture fine details of objects due to the difficulty of scaling a single implicit network to an entire scene. Our Local Implicit Grids are able to capture global structures as well as local details.

\subsection{Scene reconstruction from sparse points}\label{ssec:recon}
\paragraph{Task} As a final task and our main application, we apply our reconstruction method to the classic task in computer graphics to reconstruct geometries from sparse points. This is an important application since surface reconstruction from points is a crucial step in the process of digitizing the 3-dimensional world. The input to the reconstruction pipeline is the sparse point samples that we randomly sample from the surface mesh of the scene datasets. We study reconstruction performances with a varied number of input point samples and point densities.

\paragraph{Baseline} We mainly compare our method to the traditional Poisson Surface Reconstruction (PSR) method~\cite{kazhdan2006poisson,kazhdan2013screened} with a high octree depth value (depth=10) for the scene reconstruction experiment, which remains the state-of-the-art method for surface reconstruction tasks of scenes. We also compare with other classic (PSR at depth 8 and 9, Alpha Complex \cite{edelsbrunner1994threes}, Ball Pivoting \cite{bernardini1999balls}) and deep (Deep Geometric Prior \cite{williams2019deep}) reconstruction methods on one representative scenario (see 100$pts/m^2$ in Table \ref{tab:exp3_sc}) due to the high computational cost of evaluating all methods on all scenes. While various other deep learning based methods \cite{park2019deepsdf, liao2018deep, jiang2019convolutional} have been proposed for surface reconstruction from points in a similar setting, all of the deep learning based methods are object-specific, trained and tested on specific object categories in ShapeNet, with no anticipated transferability to unseen categories or scenes, as we have shown in the experiment in Sec. \ref{ssec:ae}. Furthermore, as both PSR and our method require no training/finetuning on the scene level datasets, the task is based on the premise that high-quality 3D training data is costly to acquire or unavailable for scenes. For our method, we adaptively use different part sizes for different point densities. We use 25cm (1000 pts/$m^2$), 35cm (500 pts/$m^2$), 50cm (100 pts/$m^2$) and 75cm (20 pts/$m^2$) corresponding to different point densities for optimal performance.

\paragraph{Data} We evaluate the reconstruction performance of the methods on a synthetic dataset: SceneNet \cite{handa2016understanding}, and a high quality scanned dataset: Matterport 3D \cite{Matterport3D} (validation split). As both SceneNet and Matterport 3D datasets are not watertight, and in addition to that, SceneNet dataset has various artifacts such as double-sided faces that produce conflicting normal samples, we preprocess both datasets using the watertight manifold algorithm as describe in \cite{huang2018robust}. For both datasets, as the scenes vary in sizes, we sample a constant density of points on mesh surfaces (20, 100, 500 and 1000 points per $m^2$). As preprocessing produces large empty volumes for SceneNet, we drop scenes that have a volume-to-surface-area ratio lower than 0.13.

\paragraph{Results Discussion} We compare the reconstruction performances in Table \ref{tab:exp3_sc} and \ref{tab:exp3_mp}, and Fig. \ref{fig:exp3_sc}. With a high number of input point samples, both PSR10 and our method are able to reconstruct the original scene with high fidelity. However, with a low number of point samples, our method is able to leverage geometric priors to perform a much better reconstruction than PSR. Additionally, our method is able to reconstruct thin structures very well whereas PSR fails to do so. However, since our method only reconstructs finite thickness surfaces as determined by finite part size, it creates double sided surfaces on the enclosed non-visible interiors, leading to degraded performance in F-Score for the 500 and 1000 pts/$m^2$ scenarios in Table \ref{tab:exp3_sc}.
\begin{table}[]
\small
\begin{tabular}{@{}lllll@{}}
\toprule
points/$m^2$                               & Method & CD($\downarrow$)& Normal($\uparrow$) & F-Score($\uparrow$) \\ \midrule
\multicolumn{1}{l|}{\multirow{2}{*}{20}}   & PSR10  & 0.077 & 0.802 & 0.317  \\
\multicolumn{1}{l|}{}                      & Ours   & \textbf{0.017} & \textbf{0.920} & \textbf{0.859}  \\ \midrule
\multicolumn{1}{l|}{\multirow{4}{*}{100}}  & PSR8   & 0.031 & 0.891 & 0.721  \\
\multicolumn{1}{l|}{}                      & PSR9   & 0.035 & 0.890 & 0.721  \\
\multicolumn{1}{l|}{}                      & PSR10  & 0.035 & 0.890 & 0.725  \\
\multicolumn{1}{l|}{}                      & Alpha  & 0.021 & 0.709 & 0.736  \\
\multicolumn{1}{l|}{}                      & BallPvt& 0.015 & 0.880 & 0.839  \\
\multicolumn{1}{l|}{}                      & DGP    & 0.037 & 0.852 & 0.571  \\
\multicolumn{1}{l|}{}                      & Ours   & \textbf{0.012} & \textbf{0.961} & \textbf{0.957}  \\ \midrule
\multicolumn{1}{l|}{\multirow{2}{*}{500}}  & PSR10  & 0.024 & 0.959 & 0.957  \\
\multicolumn{1}{l|}{}                      & Ours   & \textbf{0.010} & \textbf{0.976} & \textbf{0.972}  \\ \midrule
\multicolumn{1}{l|}{\multirow{2}{*}{1000}} & PSR10  & 0.026 & 0.975 & 0.984  \\
\multicolumn{1}{l|}{}                      & Ours   & \textbf{0.009} & \textbf{0.984} & \textbf{0.986}  \\ \bottomrule
\end{tabular}
\caption{Reconstruction performance on SceneNet dataset.}\label{tab:exp3_sc}
\end{table}

\begin{table}[]
\small
\begin{tabular}{@{}lllll@{}}
\toprule
points/$m^2$                               & Method & CD($\downarrow$)& Normal($\uparrow$) & F-Score($\uparrow$) \\
\midrule
\multicolumn{1}{l|}{\multirow{2}{*}{20}}   & PSR10  & 0.167 & 0.655 & 0.276  \\
\multicolumn{1}{l|}{}                      & Ours   & \textbf{0.028} & \textbf{0.813} & \textbf{0.691}  \\ \midrule
\multicolumn{1}{l|}{\multirow{2}{*}{100}}  & PSR10  & 0.106 & 0.757 & 0.455  \\
\multicolumn{1}{l|}{}                      & Ours   & \textbf{0.013} & \textbf{0.883} & \textbf{0.889}  \\ \midrule
\multicolumn{1}{l|}{\multirow{2}{*}{500}}  & PSR10  & 0.103 & 0.871 & 0.778  \\
\multicolumn{1}{l|}{}                      & Ours   & \textbf{0.008} & \textbf{0.928} & \textbf{0.970}  \\ \midrule
\multicolumn{1}{l|}{\multirow{2}{*}{1000}} & PSR10  & 0.102 & 0.910 & 0.862  \\
\multicolumn{1}{l|}{}                      & Ours   & \textbf{0.007} & \textbf{0.945} & \textbf{0.985}  \\ \bottomrule
\end{tabular}
\caption{Reconstruction performance on Matterport dataset.}\label{tab:exp3_mp}
\end{table}

\section{Ablation Study}
Additionally, we study the effects of two important aspects of our method: the part scale that we choose for reconstructing each scene, and overlapping latent grids. We choose SceneNet reconstruction from 100 point samples / $m^2$ as a representative case for the ablation study. See Table \ref{tab:ablate} for a comparison. As seen from the results, the reconstruction results are affected by the choice of part scale, albeit not very heavily influenced. Overlapping latent grids significantly improves the quality of the overall reconstruction. With a smaller latent code size of 8, the performance is slightly deteriorated due to more limited expressivity for part geometries.

\begin{table}[]
\small
\resizebox{\columnwidth}{!}{
\begin{tabular}{@{}llllll@{}}
\toprule
CL & PS & Overlap & CD($\downarrow$) & Normal($\uparrow$) & F-Score($\uparrow$) \\ \midrule
32 & 25cm       & Yes     & 0.013           & 0.948             & 0.921              \\
32 & 50cm       & Yes     & 0.012           & 0.961             & 0.957              \\
32 & 75cm       & Yes     & 0.013           & 0.945             & 0.929              \\
\midrule
\midrule
32 & 50cm       & No      & 0.023           & 0.886             & 0.857              \\
\midrule
\midrule
8  & 50cm       & Yes      & 0.016 &	0.925	& 0.879              \\
\bottomrule
\end{tabular}
}
\caption{Ablation study on the effects of the choice of latent code length (CL), part scale (PS), and overlapping latent grid design on the reconstruction performance for scenes.}\label{tab:ablate}
\end{table}

\section{Discussion and Future Work}
The Local Implicit Grid (LIG) representation for 3D scenes is a regular grid of overlapping part-sized local regions, each encoded with an implicit feature vector. Experiments show that LIG is capable of reconstructing 3D surfaces of objects from classes unseen in training. Furthermore, to our knowledge, it is the first learned 3D representation for reconstructing scenes from sparse point sets in a scalable manner. Topics for future work include ways to constrain the LIG optimization to produce latent codes near training examples, explore alternate implicit function representations (e.g., OccNet), and to investigate the best ways to use LIG for 3D reconstruction from image(s).

\newpage

\section*{Acknowledgements}
We would like to thank Kyle Genova, Fangyin Wei, Abhijit Kundu, Alireza Fathi, Caroline Pantofaru, David Ross, Yue Wang, Mahyar Najibi and Chris Bregler for helpful discussions, Angela Dai for help with supplemental video, JP Lewis for offering help in paper review, as well as anonymous reviewers for helpful feedback.
This work was supported by the ERC Starting Grant \textit{Scan2CAD} (804724).

{\small
\bibliographystyle{ieee_fullname}
\bibliography{ref}
}

\newpage
\onecolumn
\renewcommand\thesection{\Alph{section}}
\noindent {\Large Appendix}
\setcounter{section}{0}

\section{Additional implementation details}
\subsection{Model architecture}
\begin{figure}[ht!]
    \centering
    \includegraphics[width=.8\textwidth,trim={0 0em 0 8em},clip]{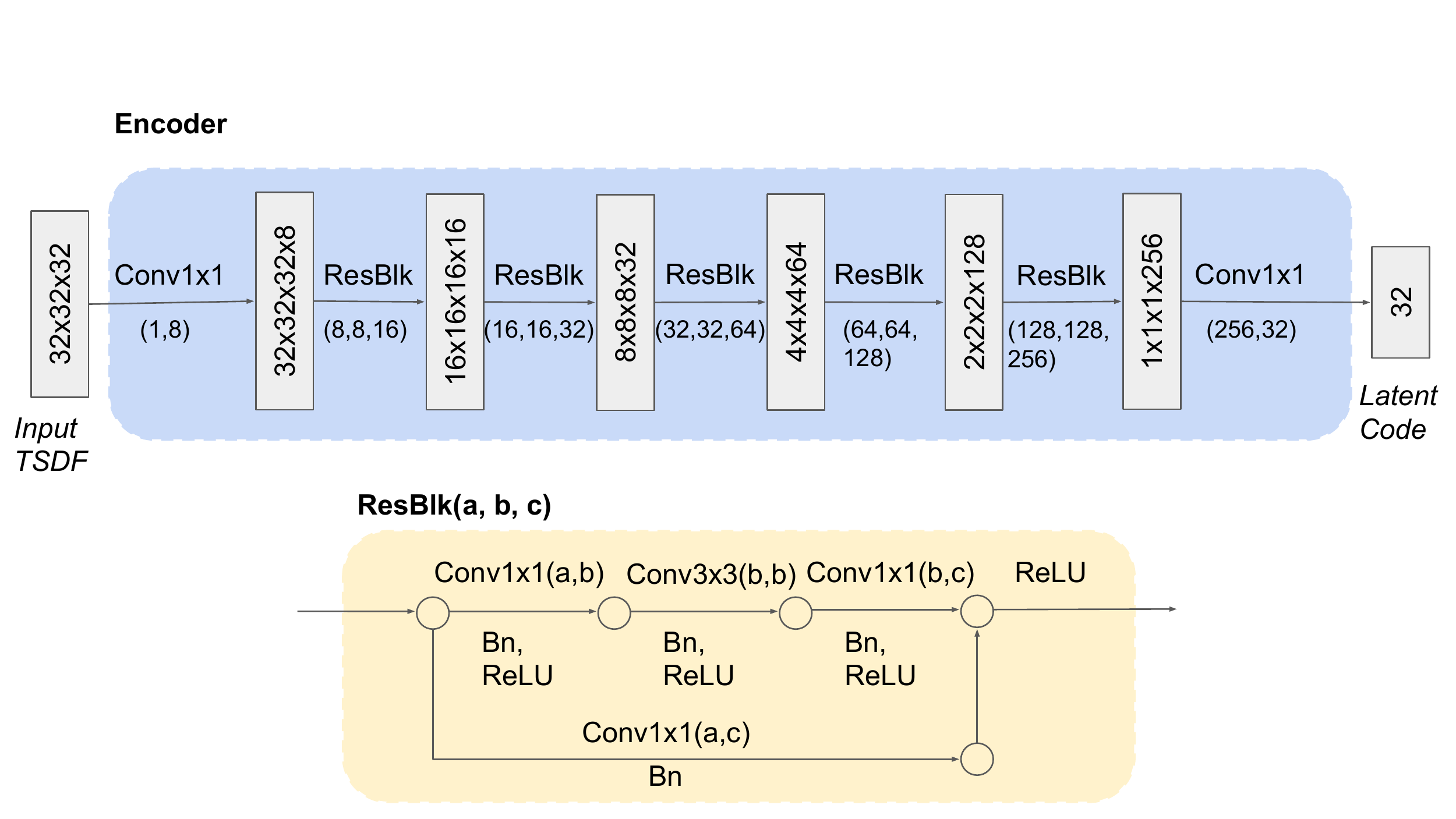}
    \caption{Encoder architecture. The encoder is a simple 3D CNN decorated with residue blocks, that encodes 3D TSDF tensors into latent codes, which can be decoded into implicit surfaces by an implicit network decoder.}\label{fig:arch}
    \vspace{-1em}
\end{figure}

We present a schematic of our encoder architecture for our part autoencoder in Fig. \ref{fig:arch}. The input to the encoder is a normalized TSDF crop of the part to be encoded, and the encoder uses 3D CNNs to encode the input into a latent code of dimensions 32. The encoder is decorated with residue blocks with bottleneck layers for improved performance.

We refer the reader to \cite{chen2019learning} for the architecture for our refiner. We preserve the architecture of the IM-NET model, but reduce the latent dimension from 128 to 32, and reduce the number of hidden layers in every layer of the model to 1/4 of the original value for improved efficiency, due to the fact that part geometries are easier to learn and represent than entire objects.
\subsection{Part autoencoder training}
For training the part autoencoder, we use a batch size of 32, and for each shape we sample 2048 point samples. We train with a latent penalty factor $\lambda=10^{-2}$, learning rate of $10^{-3}$. We sample empty volumes with a probability of $10^{-3}$ to embed empty space. We train the part autoencoder for a total of $10^{7}$ steps.

\subsection{Inference}
For reconstructing geometries from point samples, for each point sample, we sample 10 points along the point normal with a standard deviation of 1cm. For the Local Implicit Grid, we initialize each cell with Gaussian normal random values with a standard deviation of 0.01. During latent grid optimization, we use 32768 random point samples per batch, and optimize with a learning rate of $10^{-3}$. We optimize for a fixed 10000 steps. When extracting the final mesh, we extract the mesh at $1/64 m$ resolution.

\subsection{Postprocessing algorithm}
\begin{figure*}[h!]
    \centering
    \begin{subfigure}[t]{0.33\textwidth}
        \centering
        \includegraphics[width=\linewidth]{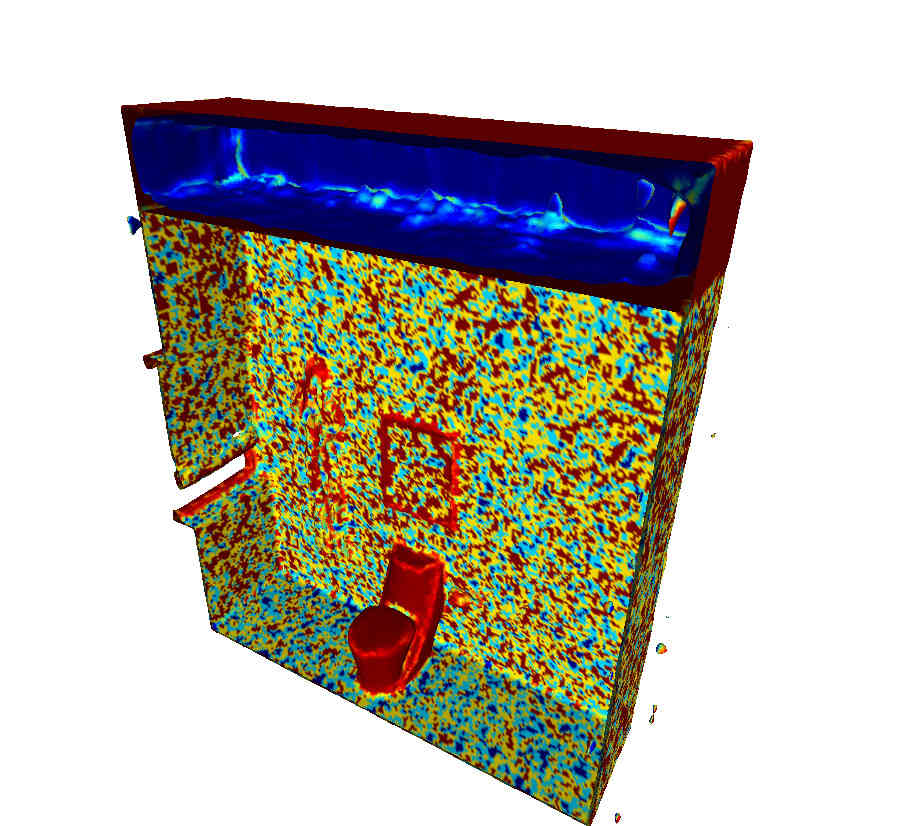}
        \caption{Before postprocessing. Color by original mesh normal alignment signal.}\label{fig:bef_nolap}
    \end{subfigure}%
    ~
    \begin{subfigure}[t]{0.33\textwidth}
        \centering
        \includegraphics[width=\linewidth]{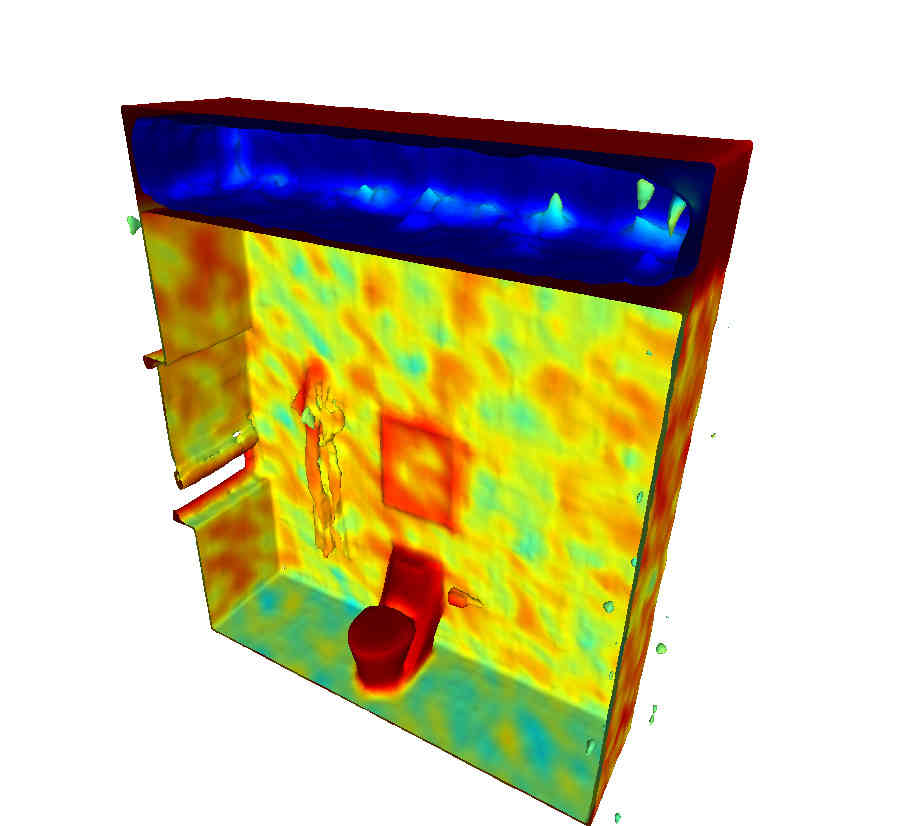}
        \caption{Before postprocessing. Color by normal alignment signal after Lap. smoothing.}\label{fig:bef_lap}
    \end{subfigure}%
    ~
    \begin{subfigure}[t]{0.33\textwidth}
        \centering
        \includegraphics[width=\linewidth]{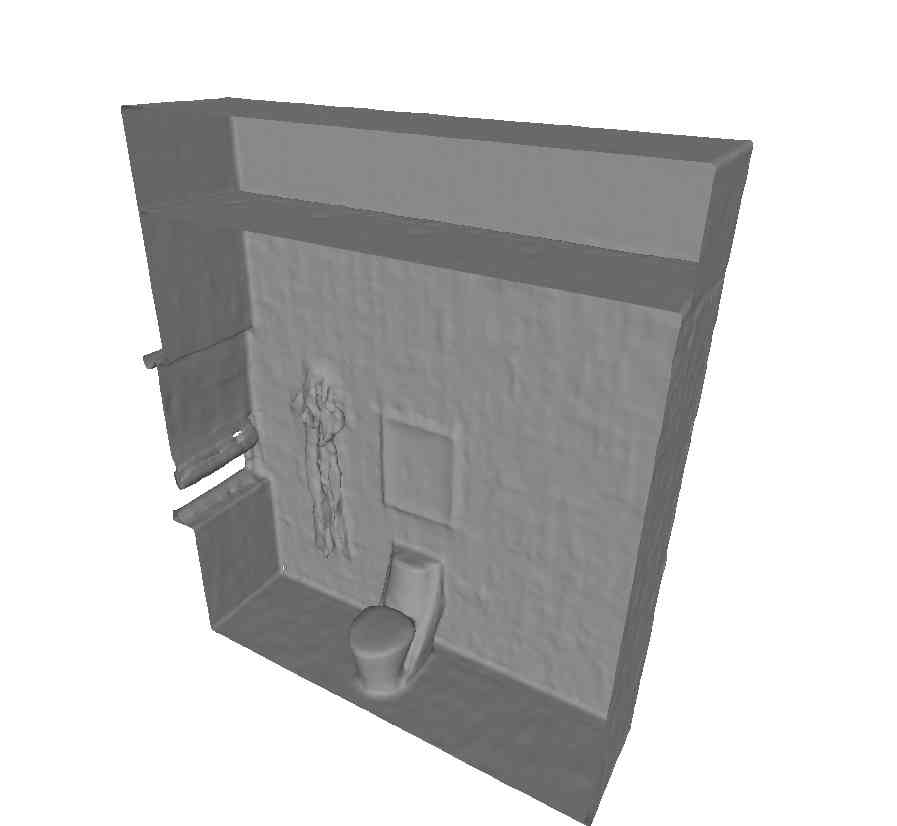}
        \caption{Postprocessed Reconstructed Mesh}\label{fig:after}
    \end{subfigure}
    \caption{Schematics for postprocessing algorithm. The back-face artifact in the original reconstructed mesh can be clearly seen in dark blue, and is effectively removed in the postprocessed mesh (c).}\label{fig:postproc}
    \vspace{-1em}
\end{figure*}

As discussed in the main text, one undesired side product from assuming all empty LIG grid cells to be ``exterior" space is that it results in  back-faces enclosed in large volumes. A simple postprocessing algorithm can be devised to remove such artifacts. For every face in the reconstructed mesh, we first compute the centroid of each face, as well as its normal direction. For the centroid of each face, we find the top-k nearest points in the original input oriented point set and compute the dot product of the normals between the pair of points. As such, back-faces will consistently have the opposite sign, and the exterior face will have the correct sign. This, however, will be noisy and non-robust to thin surfaces (with both sides very close to each other), since approximately half of the time the faces will find an input point on the opposite side as its nearest neighbor (see Fig. \ref{fig:bef_nolap}). This can be effectively mitigated by using a Laplacian kernel (diffusion coefficient $\lambda$, $i$ iterations) to smooth the normal alignment signal, followed by discarding all faces below a certain normal alignment threshold $n$, and discarding all disconnected components with an area below $a$.

In all our cases, we used the parameters $k=3, n=-0.75, \lambda=0.5, i=50, a=1$.

\section{Additional ablation studies}
\begin{figure}[h!]
\begin{floatrow}
\capbtabbox[.33\textwidth]{%
    \begin{tabular}{@{}llll@{}}
\toprule
CL & CD($\downarrow$) & Normal($\uparrow$) & F-Score($\uparrow$) \\ \midrule
8 & 0.018           & 0.925             & 0.879              \\
16 & 0.013           & 0.944             & 0.923              \\
32 & 0.012           & 0.961             & 0.957              \\
64 & 0.012           & 0.965             & 0.963              \\
\bottomrule
\end{tabular}
}{%
  \caption{Additional ablation study on the effects of latent code length (CL). Reconstruction performance measured on SceneNet reconstruction from 100 point samples / $m^2$.}%
  \label{tab:supp_abl}
}
\ffigbox[.66\textwidth]{%
    \centering
    \begin{subfigure}{.33\linewidth}
    \centering
    \captionsetup{width=.8\linewidth}
    \includegraphics[width=\linewidth, height=1.5\linewidth]{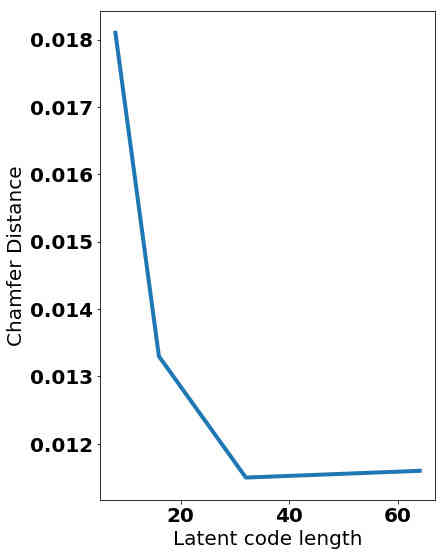}
    \caption{Chamfer Distance}
    \end{subfigure}%
    \begin{subfigure}{.33\linewidth}
    \centering
    \captionsetup{width=.8\linewidth}
    \includegraphics[width=\linewidth, height=1.5\linewidth]{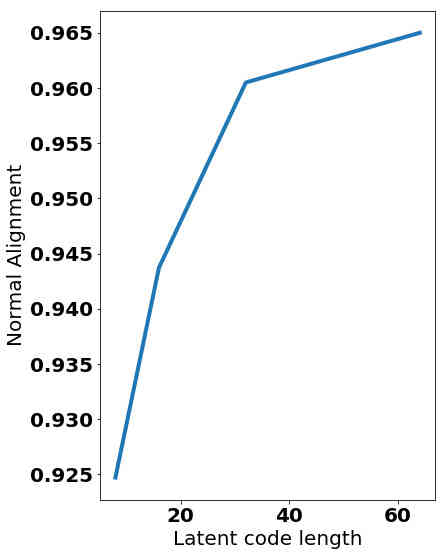}
    \caption{Normal Alignment}
    \end{subfigure}%
    \begin{subfigure}{.33\linewidth}
    \centering
    \captionsetup{width=\linewidth}
    \includegraphics[width=\linewidth, height=1.5\linewidth]{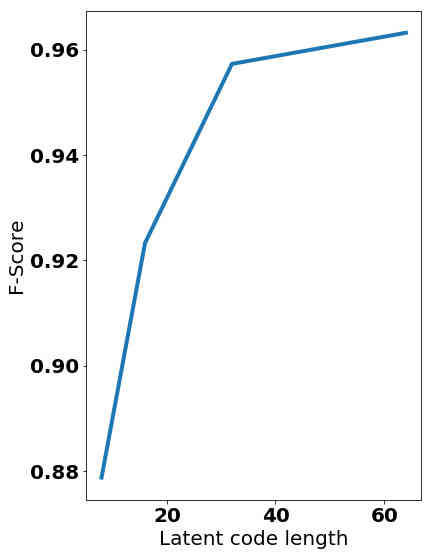}
    \caption{F-Score}
    \end{subfigure}
}{%
  \caption{Line plot for Chamfer Distance, Normal Alignment and F-Score versus Latent Code Length.}
  \label{fig:supp_abl}
}
\end{floatrow}
\end{figure}

We perform additional ablation studies on the effects of latent code length on reconstruction performace. See Table \ref{tab:supp_abl} and Fig. \ref{fig:supp_abl} for reference.  With increasing number of latent channels, the reconstruction performance improves with diminishing marginal improvement. Our choice of 32 latent channels strikes a good balance between performance and efficiency.

\newpage

\section{Additional visual results}
\begin{figure}[h!]
    \centering
    \includegraphics[width=\textwidth]{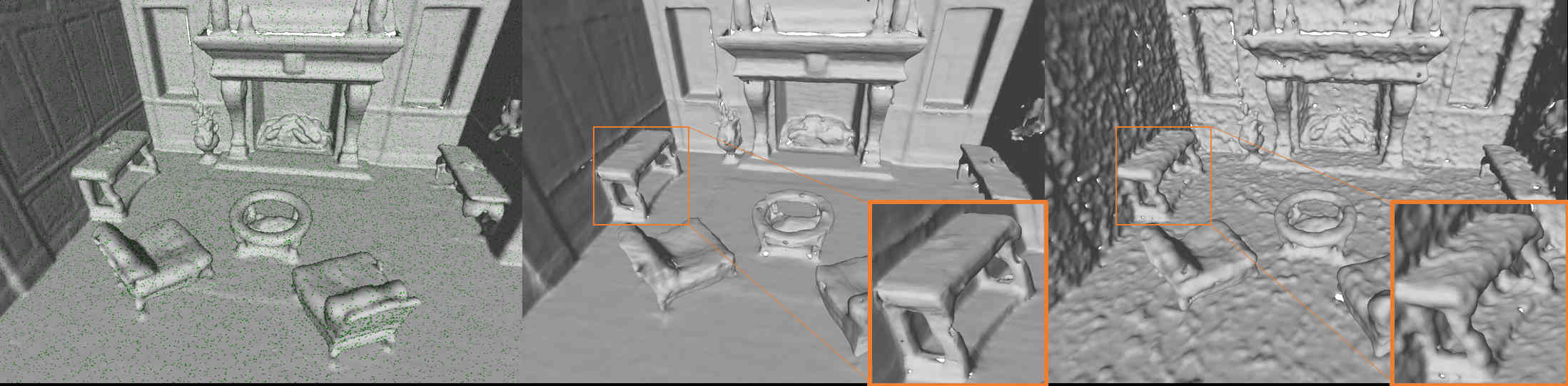}
    \includegraphics[width=\textwidth]{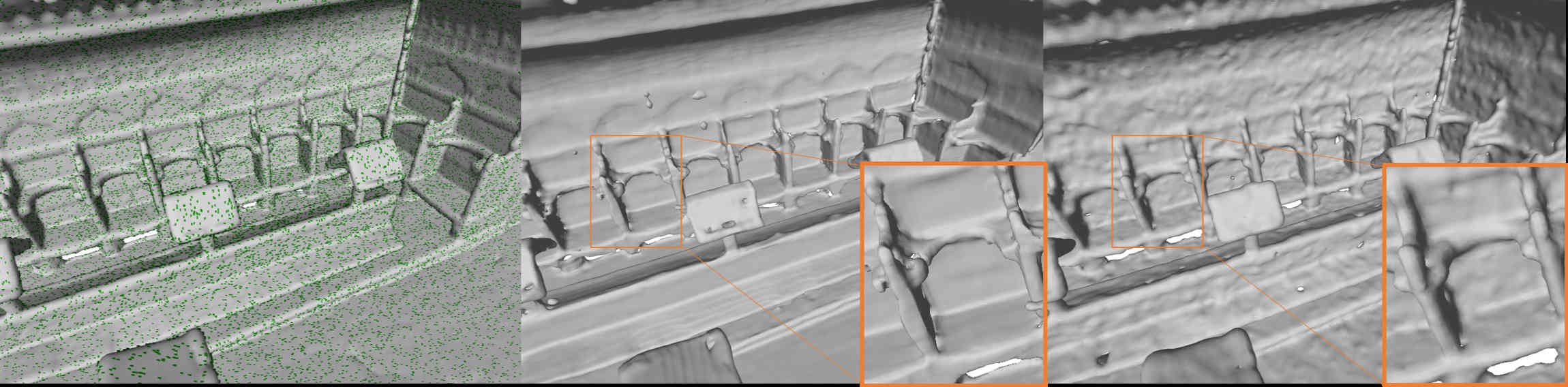}
    \includegraphics[width=\textwidth]{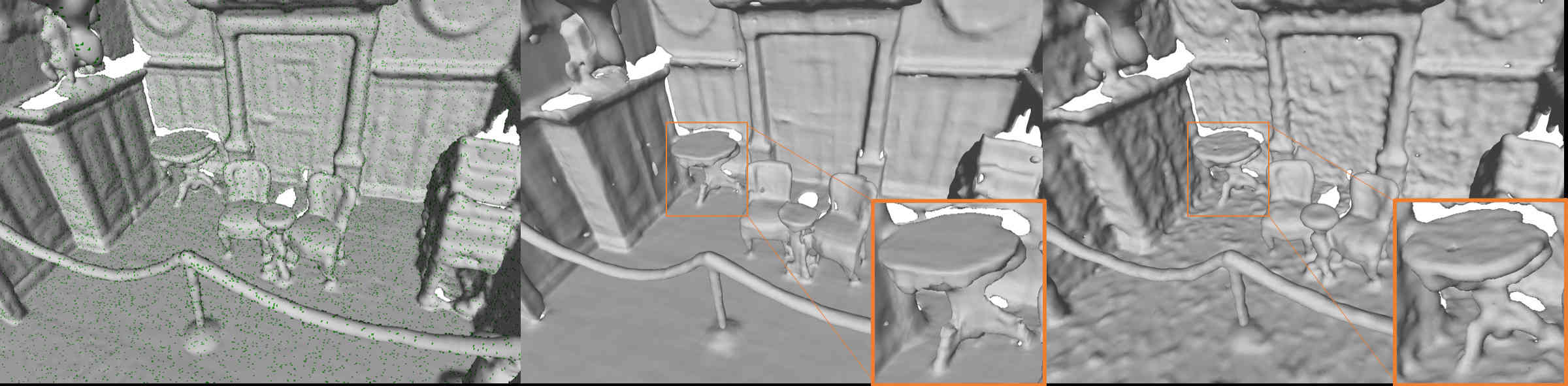}
    \includegraphics[width=\textwidth]{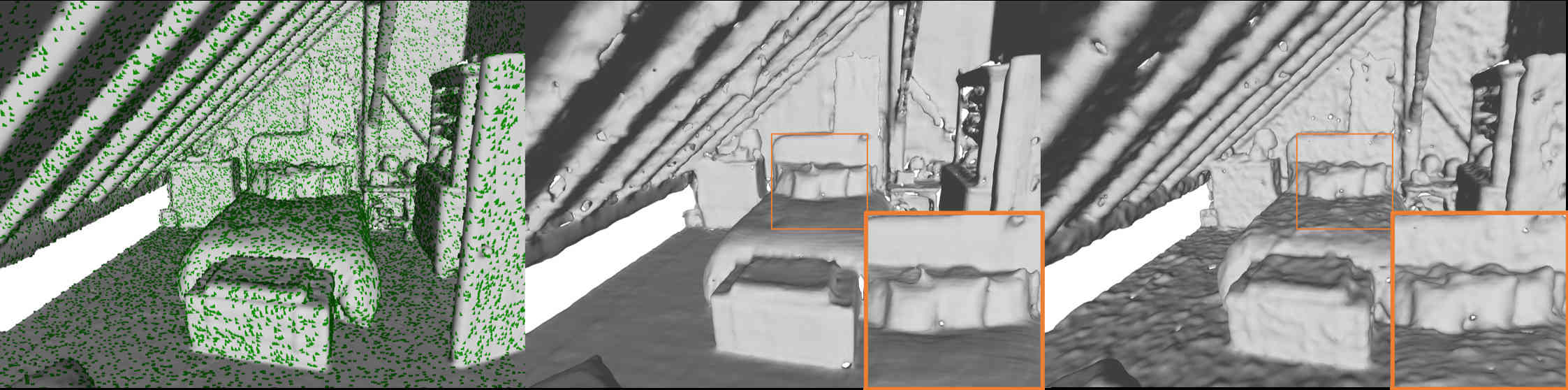}
    \caption{Left: Ground truth mesh overlaid with input point samples; Middle: Our reconstruction; Right: Screened PSR \cite{kazhdan2013screened} reconstruction. The input are point samples from the Matterport ground truth mesh at a sample density of 500 points / $m^2$.}
    \label{fig:supp_mp}
\end{figure}

\newpage

\begin{figure}[h!]
    \centering
    \includegraphics[width=\textwidth]{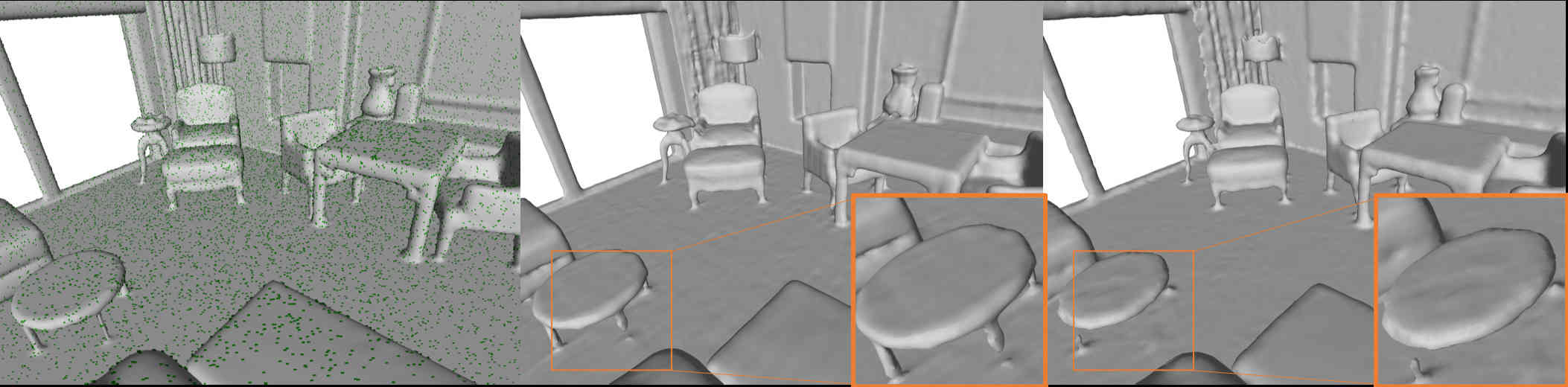}
    \includegraphics[width=\textwidth]{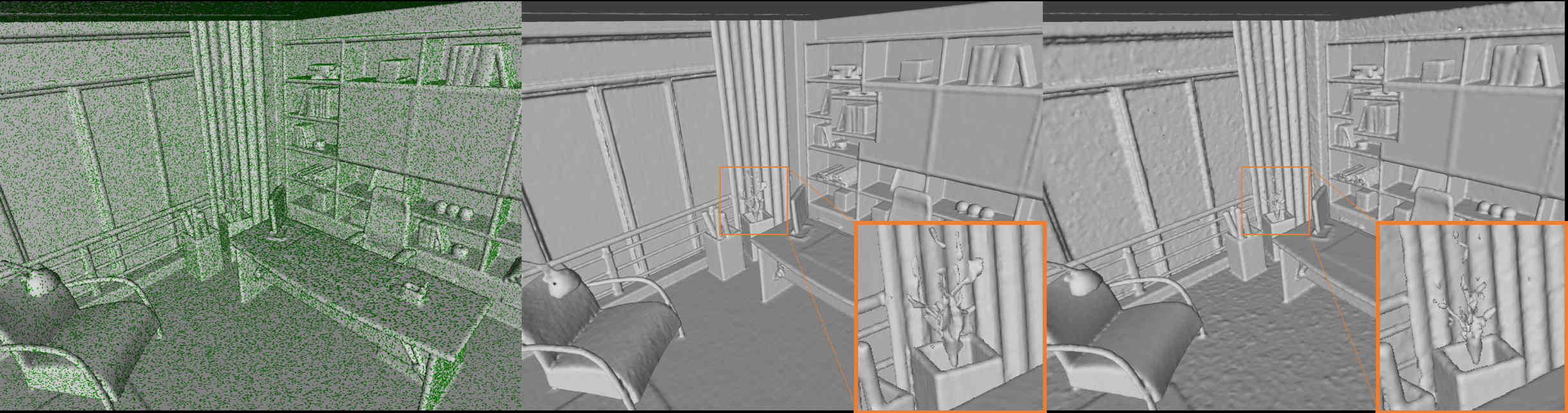}
    \includegraphics[width=\textwidth]{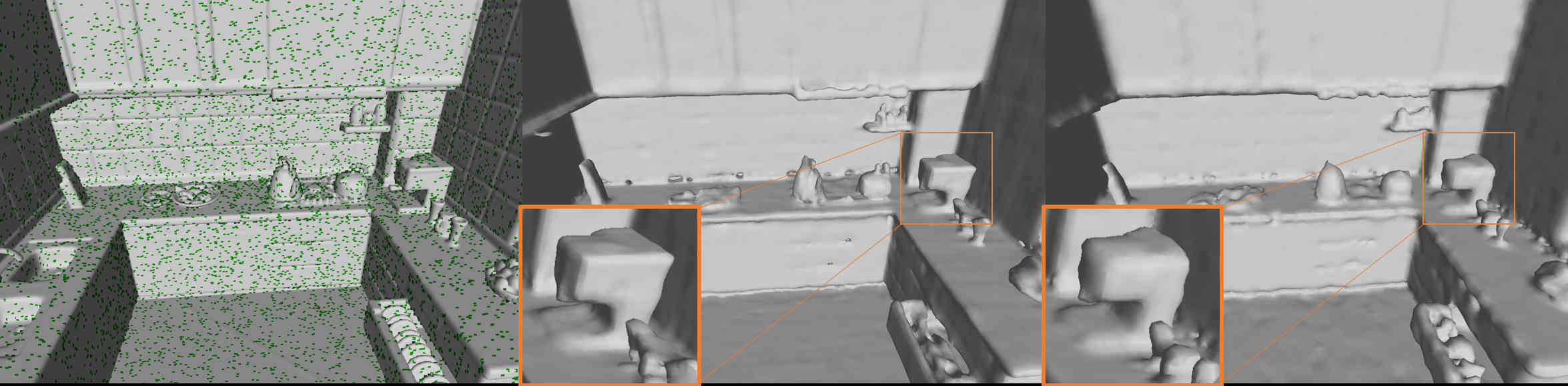}
    \includegraphics[width=\textwidth]{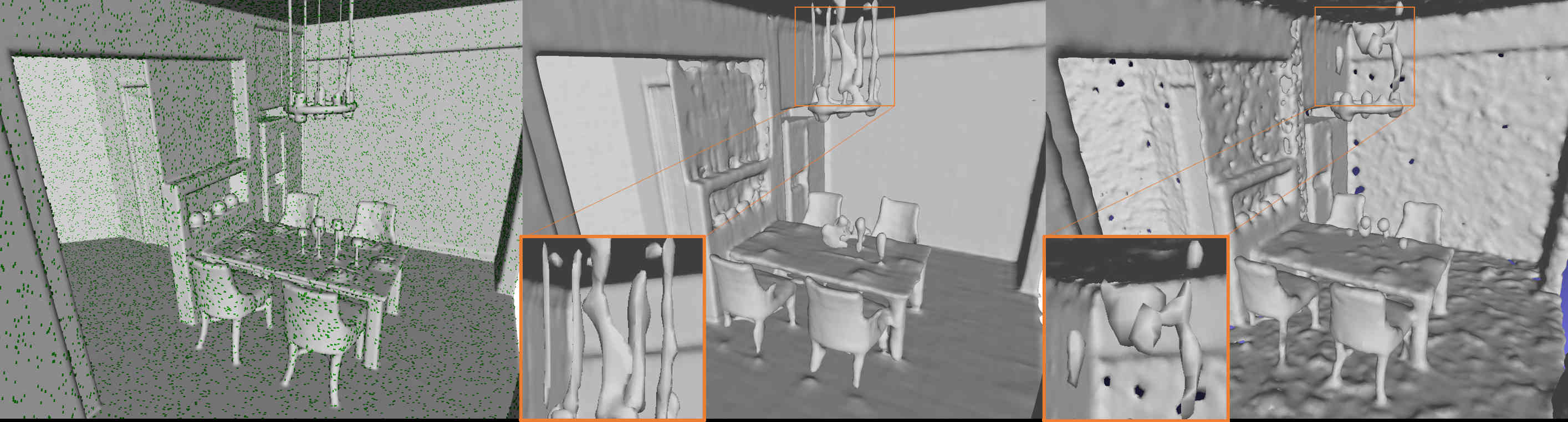}
    \caption{Left: Ground truth mesh overlaid with input point samples; Middle: Our reconstruction; Right: Screened PSR \cite{kazhdan2013screened} reconstruction. The input are point samples from the SceneNet ground truth mesh at a sample density of 500 points / $m^2$.}
    \label{fig:supp_sn}
\end{figure}

\end{document}